\documentclass[fleqn,10pt]{wlscirep}
\usepackage[utf8]{inputenc}
\usepackage[T1]{fontenc}
\title{Molecular Geometry Prediction using a Deep Generative Graph Neural Network}

\author[1]{Elman Mansimov}
\author[2]{Omar Mahmood}
\author[3]{Seokho Kang}
\author[1,2,4,5,*]{Kyunghyun Cho}
\affil[1]{Department of Computer Science, Courant Institute of Mathematical Sciences, New York University, 60 5th Avenue, New York, New York 10011, United States}
\affil[2]{Center for Data Science, New York University, 60 5th Avenue, New York, New York 10011, United States}
\affil[3]{Department of Systems Management Engineering, Sungkyunkwan University, 2066 Seobu-ro, Jangan-gu, Suwon 16419, Republic of Korea}
\affil[4]{Facebook AI Research, 770 Broadway, New York, New York 10003, United States}
\affil[5]{CIFAR Azrieli Global Scholar, Canadian Institute for Advanced Research, 661 University Avenue, Toronto, ON M5G 1M1, Canada}

\affil[*]{Correspondence should be addressed to K.C. (email: kyunghyun.cho@nyu.edu)}

\usepackage{amsmath}
\usepackage{amssymb}
\usepackage{color}
\usepackage{subcaption} 
\usepackage[utf8]{inputenc}
\usepackage[T1]{fontenc}
\usepackage{subcaption} 
\usepackage{booktabs,makecell,pbox}
\newcommand{\rulesep}{\unskip\ \vrule\ }

\def\addition#1{\textcolor{black}{#1}}

\usepackage{colortbl}

\begin{abstract}
A molecule's geometry, also known as conformation, is one of a molecule's most important properties, determining the reactions it participates in, the bonds it forms, and the interactions it has with other molecules. Conventional conformation generation methods minimize hand-designed molecular force field energy functions that are often not well correlated with the true energy function of a molecule observed in nature. They generate geometrically diverse sets of conformations, some of which are very similar to the lowest-energy conformations and others of which are very different. In this paper, we propose a conditional deep generative graph neural network that learns an energy function by directly learning to generate molecular conformations that are energetically favorable and more likely to be observed experimentally in data-driven manner. On three large-scale datasets containing small molecules, we show that our method generates a set of conformations that on average is far more likely to be close to the corresponding reference conformations than are those obtained from conventional force field methods. Our method maintains geometrical diversity by generating conformations that are not too similar to each other, and is also computationally faster. We also show that our method can be used to provide initial coordinates for conventional force field methods. On one of the evaluated datasets we show that this combination allows us to combine the best of both methods, yielding generated conformations that are on average close to reference conformations with some very similar to reference conformations.
\end{abstract}

\begin{document}
\flushbottom
\maketitle
\thispagestyle{empty}

\section*{Introduction}

The three-dimensional (3-D) coordinates of atoms in a molecule are commonly referred to as the molecule's geometry or \textbf{conformation}. The task, known as conformation generation, of predicting possible valid coordinates of a molecule, is important for determining a molecule's chemical and physical properties.\cite{conformationgen_sota} Conformation generation is also a vital part of applications such as generating 3-D quantitative structure-activity relationships (QSAR), structure-based virtual screening and pharmacophore modeling.\cite{confgenreview} Conformations can be determined in a physical setting using instrumental techniques such as X-ray crystallography as well as using experimental techniques. However, these methods are typically time-consuming and costly.

A number of computational methods have been developed for conformation generation over the past few decades.\cite{confgenreview} Typically this problem is approached by using a force field energy function to calculate a molecule's energy, and then minimizing this energy with respect to the molecule's coordinates. This hand-designed energy function yields an approximation of the molecule's true potential energy observed in nature based on the molecule's atoms, bonds and coordinates. The minimum of this energy function corresponds to the molecule's most stable configuration. Although this approach has been commonly used to generate a geometrically diverse set of conformations with certain conformations being similar to the lowest-energy conformations, it has been shown that molecule force field energy functions are often a crude approximation of actual molecular energy.\cite{lowenergyquant}

In this paper, we propose a deep generative graph neural network that learns the energy function from data in an end-to-end fashion by generating molecular conformations that are energetically favorable and more likely to be observed experimentally.\cite{ours} This is done by maximizing the likelihood of the reference conformations of the molecules in the dataset. We evaluate and compare our method with conventional molecular force field methods on three databases of small molecules by calculating the root-mean-square deviation (RMSD) between generated and reference conformations. We show that conformations generated by our model are on average far more likely to be close to the reference conformation compared to those generated by conventional force field methods \emph{i.e.} the variance of the RMSD between generated and reference conformations is lower for our method. Despite having lower variance, we show that our method does not generate geometrically similar conformations. We also show that our approach is computationally faster than force field methods.

A disadvantage of our model is that in general for a given molecule, the best conformation generated by our model lies further away from the reference conformation compared to the best conformation generated by force field methods. We show that for the QM9 small molecule dataset, the best of both methods can be combined by using the conformations generated by the deep generative graph neural network as an initialization to the force field method. 

\section*{Conformation Generation}

We consider a molecule as an undirected, complete graph $G=(V,E)$, where $V$ is a set of vertices corresponding to atoms, and $E$ is a set of edges representing the interactions between pairs of atoms from $V$. Each atom is represented as a vector $v_i \in \mathbb{R}^{d_v}$ of node features, and the edge between the $i$-th and $j$-th atoms is represented as a vector $e_{ij} \in \mathbb{R}^{d_e}$ of edge features. There are $M$ vertices and $M(M-1)/2$ edges. We define a plausible conformation as one that may correspond to a stable configuration of a molecule. Given the graph of a molecule, the task of molecular geometry prediction is the generation of a set of plausible conformations $X_a=(x_1^a, \ldots, x_M^a)$, where $x_i^a \in \mathbb{R}^3$ is a vector of the 3-D coordinates of the $i$-th atom in the $a$-th conformation. 

Molecules can transition between conformations and end up in different local minima based on the stability of the respective conformations and environmental conditions. As a result, there is more than one plausible conformation associated with each molecule; it is hence natural to formulate conformation generation as finding (local) minima of an energy function $\mathcal{F}(X, G)$ defined on a pair of molecule graph and conformation:
\begin{align}
\label{eq:energy_minimization}
    \left\{ X_1, \ldots, X_S \right\} = \arg\min_{X} \mathcal{F}(X, G).
\end{align}
Alternatively, we could sample from a Gibbs distribution:
\begin{align}
\label{eq:energy_sampling}
    \left\{ X_1, \ldots, X_S \right\} \sim p_{\mathcal{F}}(X|G),
\end{align} 
where
\begin{align}
\label{eq:gibbs}
    p_{\mathcal{F}}(X|G) = \frac{1}{\zeta(G)} \exp\left\{-\mathcal{F}(X, G)\right\},
\end{align}
where $\zeta$ is a normalizing constant. We use $S$ to indicate the number of conformations we generate for each molecule. 

Under this view, the problem of conformation generation is decomposed into two stages. In the first stage, a computationally-efficient energy function $\mathcal{F}(X, G)$ is constructed. The second stage involves either performing optimization as in Eq.~\eqref{eq:energy_minimization} or sampling as in Eq.~\eqref{eq:energy_sampling} to generate a set of conformations from this energy function. 

\subsection*{Energy Function Construction}

A conventional approach is to define an energy function semi-automatically. The functional form of an energy function is designed carefully to incorporate various chemical properties, whereas detailed parameters of the energy function are either computationally or experimentally estimated. Two examples of widely used energy functions are the Universal Force Field (UFF)\cite{UFF} and the Merck Molecular Force Field (MMFF).\cite{MMFF} In contrast to these methods, here we will describe how to estimate the energy function or probability distribution directly from data using the latest techniques from deep learning.

\subsection*{Energy Minimization/Sampling}

Once the energy function is defined, a conventional approach is to run the minimization many times starting from different initial conformations. Due to the non-convexity of the energy function, each run is likely to end up in a unique local minimum, allowing us to collect a set of many conformations. 

A typical approach is to use distance geometry (DG)\cite{distancegeom} or its variants, such as experimental-torsion basic knowledge distance geometry (ETKDG),~\cite{ETKDG} to randomly generate an initial conformation that satisfies various geometric constraints such as lower and upper bounds on the distances between atoms. Starting from the initial conformation, an iterative optimization algorithm, such as L-BFGS,~\cite{liu1989limited} gradually updates the conformation until it finds a minimum of the energy function. In this paper, we instead propose an approach based on deep generative models that allow us to sample directly from a distribution over all possible conformations given a molecule graph. 

\section*{Deep Generative Model for Molecular Geometry}

We propose to ``learn'' an energy function $\mathcal{F}(G, X)$ from a database containing many pairs of a molecule and its experimentally obtained conformation. Let $\mathcal{D}=\left\{ (G_1, X_1^*), \ldots, (G_N, X_N^*) \right\}$ be a set of examples from such a database, where $X_n^*$ is ``a'' {reference conformation, often obtained and verified empirically in a certain environment.} {These reference conformations may not necessarily correspond to the lowest energy configurations of the molecules, but are energetically favorable and more likely to be observed experimentally.} Learning an energy function can then be expressed as the following optimization problem:
\begin{align}
\label{eq:mle}
    \hat{\mathcal{F}}(G, X) 
    = 
    \arg\max_{\mathcal{F}} 
    \frac{1}{N} \sum_{n=1}^N
    \underbrace{\log p_{\mathcal{F}}(X_n^* | G_n)}_{\text{(a)}},
\end{align}
where $p_{\mathcal{F}}$ is a Gibbs distribution defined using $\mathcal{F}$ as in Eq.~\eqref{eq:gibbs}. In other words, we can learn the energy function $\mathcal{F}$ by maximizing the log-likelihood of the data $D$. \addition{In principle, the term ``energy'' has a very specific meaning in each context (e.g., potential energy, statistical free energy and etc). In our case, ``energy'' refers to an objective function that reflects the likelihood of a conformation given a molecular graph. }

\subsection*{Conditional Variational Graph Autoencoders}

We use a conditional version of a variational autoencoder~\cite{kingma2013auto} to model the distribution $p_{\mathcal{F}}$ in Eq.~\eqref{eq:mle}~(a). This choice enables an underlying model to capture the complicated, multi-modal nature of this distribution, while allowing us to efficiently sample from this  distribution. This is done by introducing a set of latent variables $Z=\left\{z_1, \ldots, z_M\right\}$, where $z_m\in \mathbb{R}^{d_z}$ and rewriting the conditional log-probability $\log p_{\mathcal{F}}(X | G)$ as
\begin{align}
\label{eq:lvm}
    \log p(X | G)
    = \log \int p (X | Z, G) p(Z | G) \text{d}Z,
\end{align}
where we omit the subscript $\mathcal{F}$ for brevity.

The marginal log-probability in Eq.~\eqref{eq:lvm} is generally intractable to compute, and we instead maximize the stochastic approximation to its lower bound, as is standard practice in problems involving variational inference:
\begin{align}
    \label{eq:lowerbound}
    \log p(X | G)
    \geq&
    \mathbb{E}_{Z \sim Q(Z|G, X)} [ 
    \log \underbrace{p(X | Z, G)}_{\text{(b) likelihood}}
    ]
    - 
    \text{KL}(\underbrace{Q(Z|G,X)}_{\text{(c) posterior}} \| \underbrace{P(Z | G)}_{\text{(a) prior}}) 
    \\
    \label{eq:sto_lowerbound}
    \approx&
    \frac{1}{K}
    \sum_{k=1}^K 
    \log p(X | Z^k, G)
    -
    \text{KL}(Q(Z|G,X) \| P(Z | G)) 
    ,
\end{align}
where $Z^k$ is the $k$-th sample from the (approximate) posterior distribution $Q$ above. We assume that we can compute the KL divergence analytically, for instance by constructing $Q$ and $P$ to be normal distributions.

\subsubsection*{Modeling the Graph using a Message Passing Neural Network}

We use a message passing neural network (MPNN)~\cite{gilmer2017neural}, a variant of a graph neural network,~\cite{scarselli2009graph,bruna2013spectral} which operates on a graph $G$ directly and is invariant to graph isomorphism. The MPNN consists of $L$ layers. At each layer $l$, we update the hidden vector $h(v_i) \in \mathbb{R}^{d_h}$ of each node and hidden matrix $h(e_{ij}) \in \mathbb{R}^{d_{h} \times d_{h}}$ of each edge using the equation
\begin{align}
\label{eq:mpnn}
    h^{l}(v_{i}) = \text{GRU}(h^{l-1}(v_{i}), J(h^{l-1}(v_{i}), h^{l-1}(v_{j \neq i}), h(e_{i,j\neq i})),
\end{align}
where $J$ is a linear one layer neural network that aggregates the information from neighboring nodes according to its hidden vectors of respective nodes and edges. $\text{GRU}$ is a gated recurrent network that combines the new aggregate information and its corresponding hidden vector from previous layer~\cite{cho2014learning}. The weights of the message passing function $J$ and $\text{GRU}$ are shared across the $L$ layers of the MPNN. 

\subsubsection*{Prior Parameterization}

We use the MPNN described above to model the prior distribution $P(Z|G)$ in Eq.~\eqref{eq:lowerbound}~(a). We initialize $h^0(v_i)$ and $h(e_{ij})$ in Eq.~\eqref{eq:mpnn} as linear transformations of the feature vectors $v_i$ and $e_{ij}$ of the nodes and edges respectively:
\begin{align}
    h^0(v_i) = U^{\text{prior}}_{\text{node}} v_i ;\quad
    h(e_{ij}) = U^{\text{prior}}_{\text{edge}} e_{ij},
\end{align}
where $U^{\text{prior}}_{\text{node}}$ and $U^{\text{prior}}_{\text{edge}}$ are matrices representing the linear transformations for the nodes and edges respectively. The final hidden vector $h^{L}(v_i)$ of each node is passed through a two layer neural network with hidden size $d_{f}$, whose output $\tilde{h}^{L}(v_i)$ is transformed into the mean and variance vectors of a Normal distribution with a diagonal covariance matrix:
\begin{align}
    \mu_i = W_{\mu}^{\text{prior}} \tilde{h}^L(v_i) + b_{\mu}^{\text{prior}}; \\
    \sigma_i^2 = \exp\left\{ W_{\sigma}^{\text{prior}} \tilde{h}^L(v_i) + b_{\sigma}^{\text{prior}} \right\},
\label{eq:prior_out}
\end{align}
where $W_{\mu}^{\text{prior}}$ and $W_{\sigma}^{\text{prior}}$ are the weight matrices and $b_{\mu}^{\text{prior}}$ and $b_{\sigma}^{\text{prior}}$ are the bias terms of the transformations.
These are used to form the prior distribution:
\begin{align}
    \label{eq:prior}
    \log P(Z|G) = \sum_{i=1}^N \sum_{j=1}^3 -\frac{(\mu_{i,j} - z_{i,j})^2}{2\sigma_{i,j}^2} - \log \sqrt{2\pi \sigma_{i,j}^2},
\end{align}
where $\mu_{i,j}$ and $\sigma^2_{i,j}$ are the $j$-th components of the mean and variance vectors respectively. In other words, we parameterize the prior distribution as a factorized Normal distribution factored over the vertices and the dimensions in the 3-D coordinate.

\subsubsection*{Likelihood Parameterization}

We use a similar MPNN to model the likelihood distribution, $P(X|Z,G)$ in Eq.~\eqref{eq:lowerbound}~(b). The only difference is that this distribution is conditioned not only on the molecular graph $G=(V,E)$ but also on the latent set $Z=\left\{ z_1, \ldots, z_M \right\}$. We incorporate the latent set $Z$ by adding the linear transformation of the node feature vector $v_i$ to its corresponding latent variable $z_i$. This result is used to initialize the hidden vector:
\begin{align}
    h^0(v_i) = U^{\text{likelihood}}_{\text{node}}v_i + z_i ;\quad h(e_{ij}) = U^{\text{likelihood}}_{\text{edge}} e_{ij},
\end{align}
where $U^{\text{likelihood}}_{\text{node}}$ and $U^{\text{likelihood}}_{\text{edge}}$ are matrices representing the linear transformations for the nodes and edges respectively. From there on, we run neural message passing as in Eqs.~(\ref{eq:mpnn}--\ref{eq:prior_out}), with a new set of parameters, $\theta_{\text{likelihood}}$, $W_{\mu}^{\text{likelihood}}$, $b_{\mu}^{\text{likelihood}}$,
$W_{\sigma}^{\text{likelihood}}$ and $b_{\sigma}^{\text{likelihood}}$. The final mean and variance vectors are now three dimensional, representing the 3-D coordinates of each atom, and we can compute the log-probability of the coordinates using Eq.~\eqref{eq:prior}.

\subsubsection*{Posterior Parameterization}

As computing the exact posterior $P(Z|G,X)$ is intractable, we resort to amortized inference using a parameterized, approximate posterior $Q(Z|G,X)$ in Eq.~\eqref{eq:lowerbound}~(c). We use a similar approach to our parameterization of the prior distribution above. However, we replace the input to the MPNN with the concatenation of an edge feature vector $e_{ij}$ and the corresponding distance (proximity) matrix $D(X^*)$ of the reference 3-D conformation $X^*$:
\begin{align}
    h(e_{ij}) = U^{\text{posterior}}_{\text{edge}} \left[ 
    \begin{array}{c}
    e_{ij} \\
    D(x^*_i)
    \end{array}
    \right].
\end{align}
With a new set of parameters, $\theta_{\text{posterior}}$, $W_{\mu}^{\text{posterior}}$, $b_{\mu}^{\text{posterior}}$,
$W_{\sigma}^{\text{posterior}}$ and $b_{\sigma}^{\text{posterior}}$, the MPNN outputs a Normal distribution for each latent variable $z_i$. Linear weight embeddings of nodes $U_{\text{node}}$ are shared between prior, likelihood and posterior.

\subsection*{Training the Conditional Variational Graph Autoencoder}

With the choice of the Gaussian latent variables $z_i$, we can use the reparameterization trick~\cite{kingma2013auto} to compute the gradient of the stochastic approximation to the lower bound in Eq.~\eqref{eq:sto_lowerbound} with respect to all the parameters of the three distributions.\cite{kingma2013auto} This property allows us to train this model on a large dataset using stochastic gradient descent (SGD). However, there are two major considerations that must be made before training this model on a large molecule database.

\subsubsection*{Post-Alignment Likelihood}

An important property of conformation generation over a usual problem of regression is that we must take into account rotation and translation. Let $R$ be an alignment function that takes as input a target conformation and a predicted conformation. The function aligns the reference conformation to the predicted conformation and returns the aligned reference conformation. $\hat{X}=R(X, X^*)$ is the conformation obtained by rotating and translating the reference conformation $X^*$ to have the smallest distance to the predicted conformation $X$ according to a predefined metric such as RMSD:
\begin{align}
    \label{eq:rmsd}
    \text{RMSD}(\hat{X}, X^*) = \sqrt{
    \frac{1}{M} \sum_{i=1}^M \| \hat{x}_i - x^*_i \|^2
    }.
\end{align}
This alignment function $R$ is selected according to the problem at hand, and we present below its use in a general form without exact specification.

We implement this invariance to rotation and translation by parameterizing the output of the likelihood distribution above to be aligned to the target molecule. That is,
\begin{align}
    \label{eq:likelihood}
    \log p(X|G, Z) = \sum_{i=1}^M \sum_{j=1}^3 -\frac{(\mu_{i,j} - \Hat{x}^*_{i,j})^2}{2\sigma_{i,j}^2} - \log \sqrt{2\pi \sigma_{i,j}^2},
\end{align}
where $\Hat{x}^*_{i}$ is the coordinate of the $i$-th atom aligned to the mean conformation $\left\{ \mu_1, \ldots, \mu_N \right\}$. That is,
\begin{align}
    \left\{ \Hat{x}^*_1, \ldots, \Hat{x}^*_M \right\}
    = R( \left\{ \mu_1, \ldots, \mu_M \right\}, X^* ).
\end{align}

In other words, we rotate and translate the reference conformation $X^*$ to be best aligned to the predicted conformation (or its mean) before computing the log-probability. This encourages the model to assign high probability to a conformation that is easily aligned to the reference conformation $X^*$, which is precisely the goal of maximum log-likelihood.

\subsubsection*{Unconditional Prior Regularization}

The second term in the lower bound in Eq.~\eqref{eq:lowerbound}, which is the KL divergence between the approximate posterior and prior, does not have a point minimum but an infinitely long valley consisting of minimum values. Consider the KL divergence between two univariate Normal distributions:
\begin{align}
    \text{KL}(\mathcal{N}(\mu_1, \sigma_1^2) \| \mathcal{N}(\mu_2, \sigma_2^2))
    =
    \log \frac{\sigma_2}{\sigma_1} +
    \frac{\sigma_1^2 + (\mu_1 - \mu_2)^2}{2 \sigma_2^2}
    -\frac{1}{2}.
\end{align}
When both distributions are shifted by the same amount, the KL divergence remains unchanged. This could lead to a difficulty in optimization, as the means of the posterior and prior distributions could both diverge. 

In order to prevent this pathological behavior, we introduce an unconditional prior distribution $P(Z)$ which is a factorized Normal distribution:
\begin{align}
    \label{eq:uncond_prior}
    P(Z) = \prod_{i=1}^M \mathcal{N}(z_i | 0, I),
\end{align}
where $\mathcal{N}$ computes a Normal probability density, and $I$ is a $d_z \times d_z$ identity matrix. We minimize the KL divergence between the original prior distribution $P(Z|G)$ and this unconditional prior distribution $P(Z)$ in addition to maximizing the lowerbound, leading to the following final objective function for each molecule:
\begin{align}
\label{eq:final_loss}
    \mathcal{L} =
    \log p(X | Z^1, G)
    -
    \text{KL}(Q(Z|G,X) \| P(Z | G)) 
    -
    \alpha \cdot \text{KL}(P(Z|G) \| P(Z)),
\end{align}
where we assume $K=1$ and introduce a coefficient $\alpha \geq 0$.

\subsection*{Inference: Predicting Molecular Geometry}

Learning a conditional variational autoencoder above corresponds to the first stage of conformation generation, that is, the stage of energy function construction. Once the energy function is constructed, we need to sample multiple conformations from the Gibbs distribution defined using the energy function, which is $\log P(X|G)$ in Eq.~\eqref{eq:lvm}. Our parameterization of the Gibbs distribution using a directed graphical model~\cite{pearl1986fusion} allows us to efficiently sample from this distribution. We first sample from the prior distribution, $\tilde{Z} \sim P(Z|G)$, and then sample from the likelihood distribution, $\tilde{X} \sim P(X | \tilde{Z}, G)$. In practice, we fix the output variance $\sigma_{i,j}$ of the likelihood distribution to be $1$ and take the mean set $\left\{ \mu_1, \ldots, \mu_M \right\}$ as a sample from the model. 

\section*{Experimental Setup}

\subsection*{Data}

\begin{figure}[t]
    \begin{minipage}{0.48\textwidth}
    \centering
        \includegraphics[width=\textwidth]{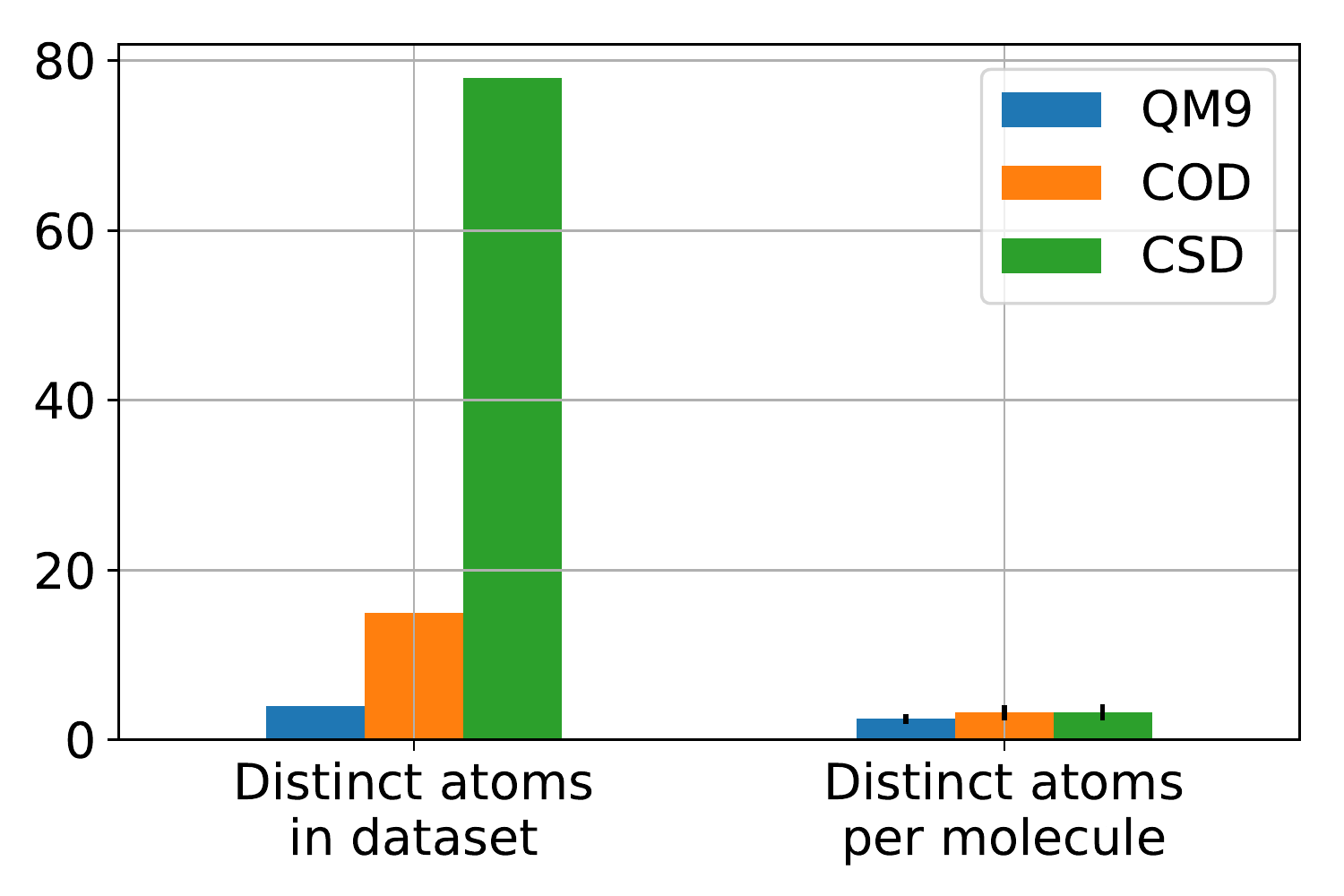}
        
        (a) Diversity of atoms overall and per molecule
    \end{minipage}
    \begin{minipage}{0.48\textwidth}
    \centering
        \includegraphics[width=\textwidth]{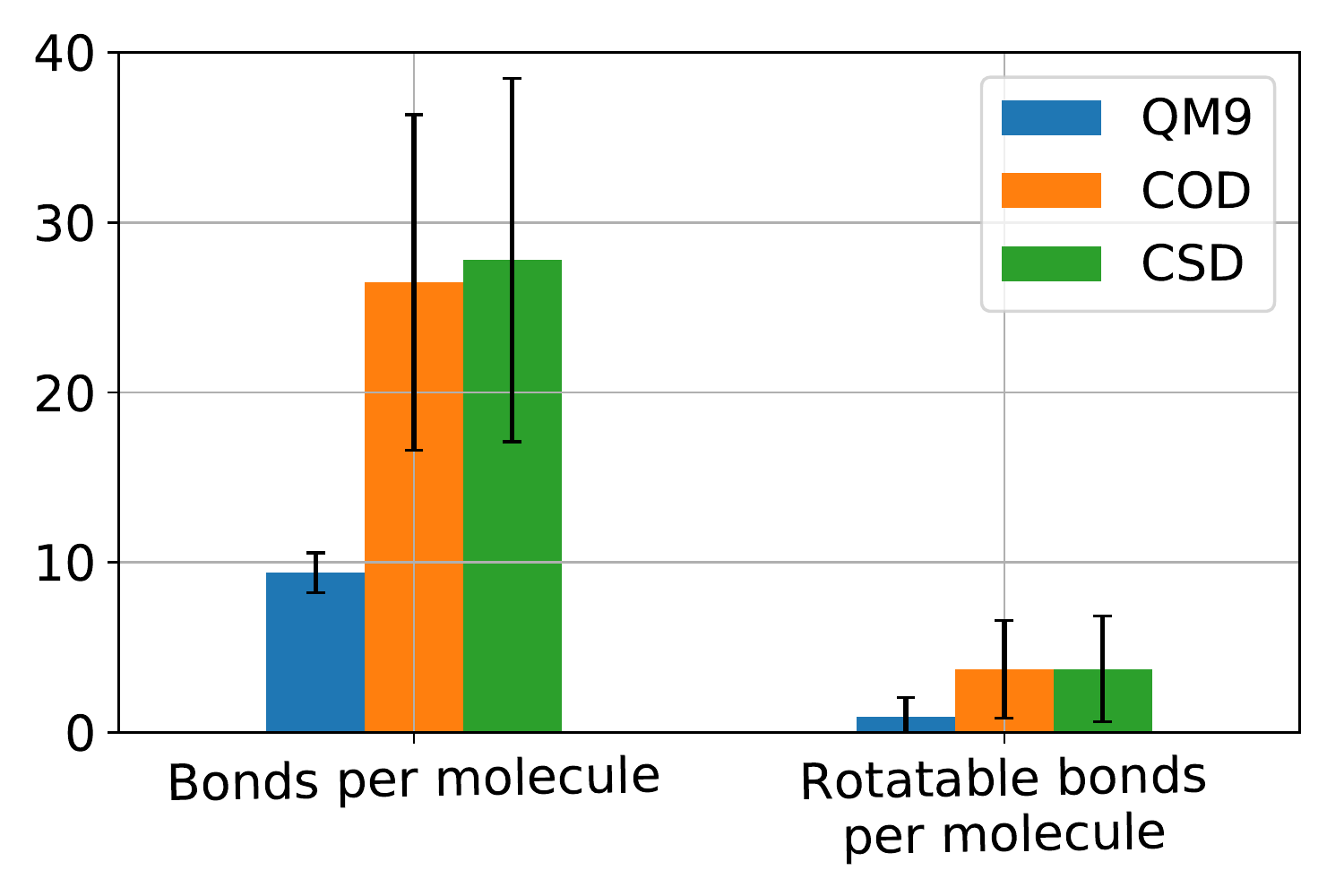}
        
        (b) Number and rotatability of bonds per molecule
    \end{minipage}
    
    \begin{minipage}{0.48\textwidth}
    \centering
        \includegraphics[width=\textwidth]{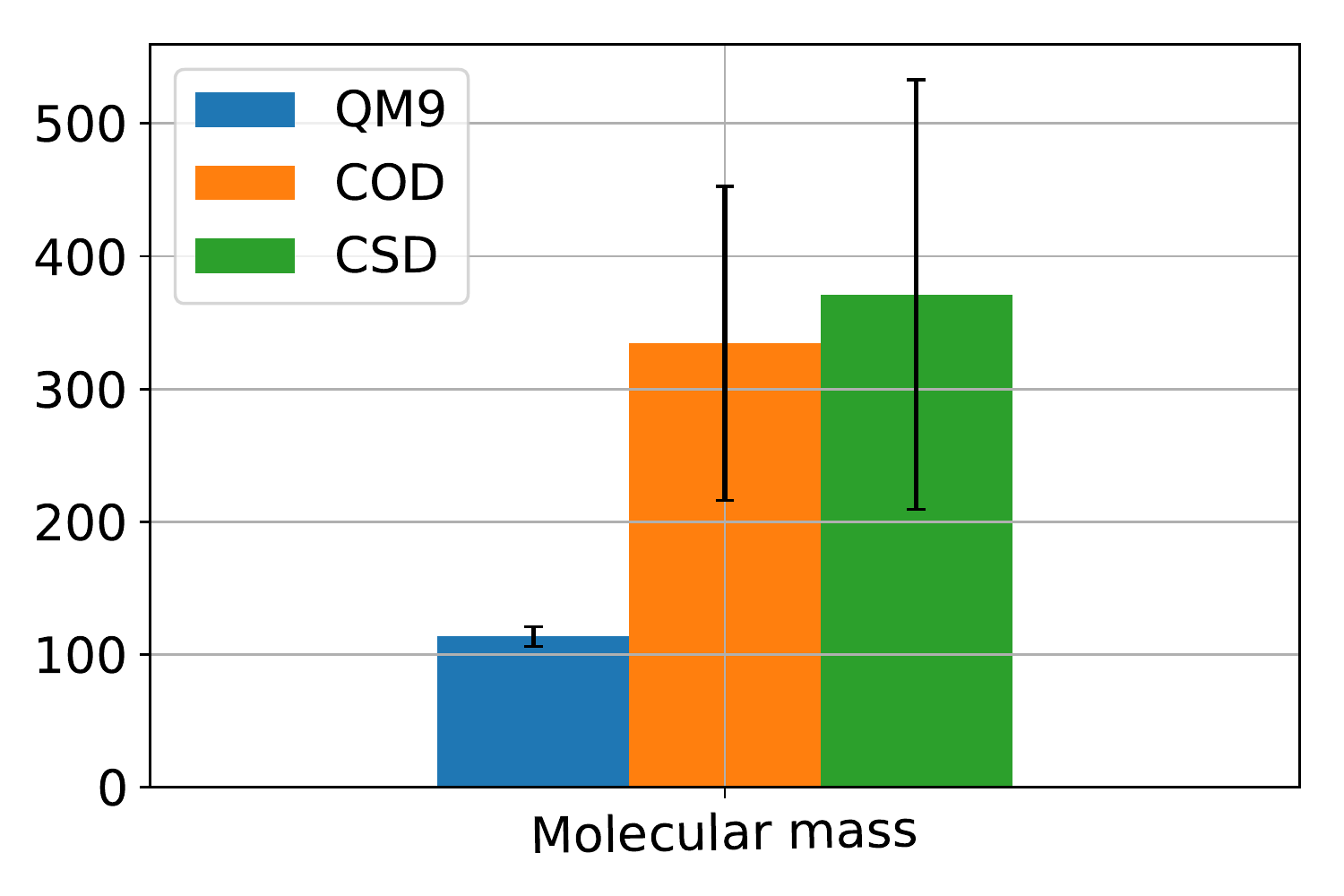}
        
        (c) Molecular mass per molecule
    \end{minipage}
    \begin{minipage}{0.48\textwidth}
    \centering
        \includegraphics[width=\textwidth]{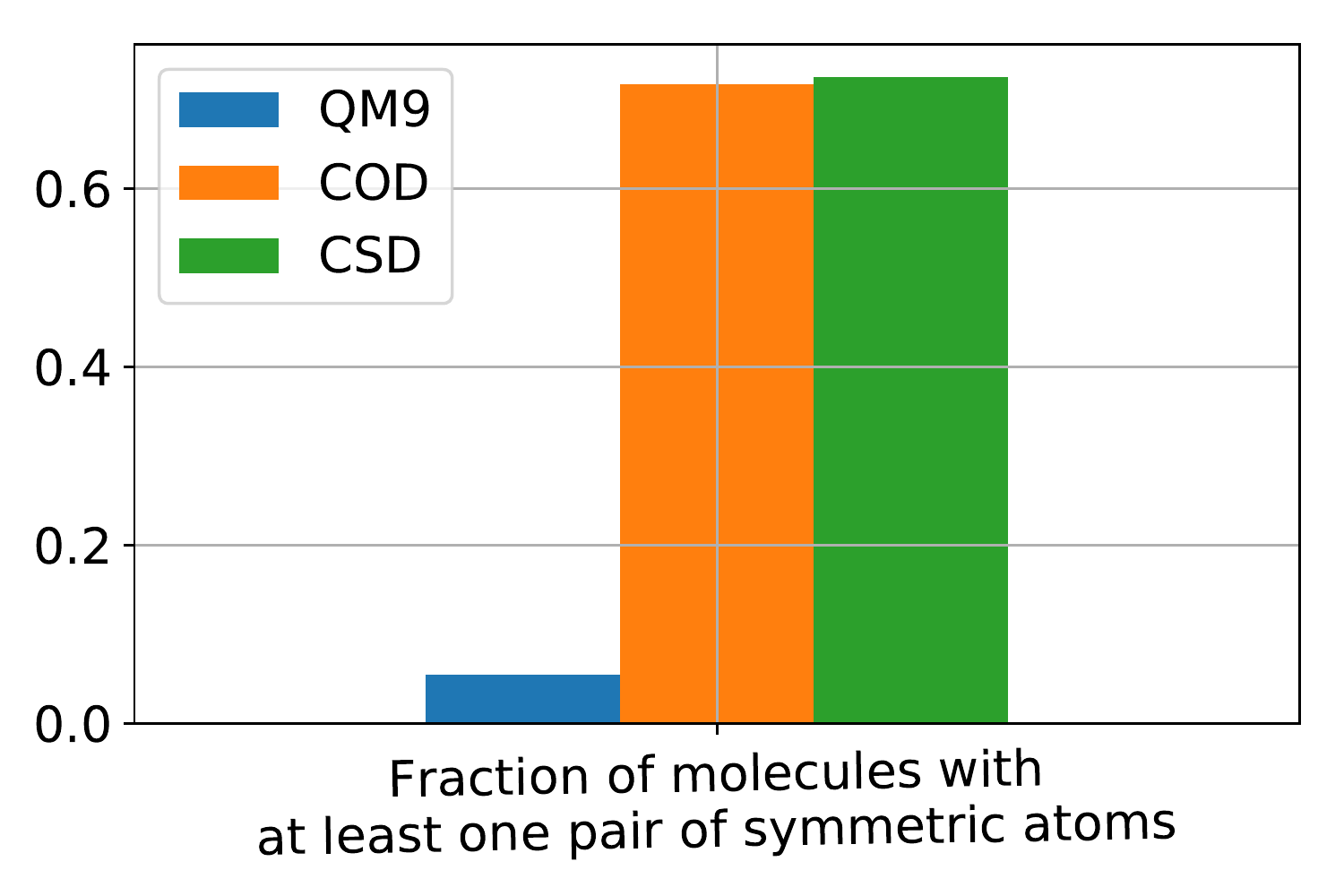}
        
        (d) Proportion of molecules with symmetry
    \end{minipage}
    \caption{Dataset Characteristics: information regarding the atoms, bonds, molecular mass and symmetry of molecules in each dataset.}
    \label{dataset_bars}
\end{figure}

We experimentally verify the effectiveness of the proposed approach using three databases of molecules: QM9,\cite{QM91,QM92} COD\cite{COD} and CSD.\cite{CSD} These datasets are selected as they possess distinct properties from each other, which allows us to carefully study various aspects of the proposed approach. There is an overlap between COD and CSD databases, since both of these databases were based on published crystallography data. We only keep molecules from each database that can be processed by RDKit\footnote{
Version 2018.09.1
}. We further remove disconnected compounds \emph{i.e.} those whose Simplified Molecular-Input Line-Entry System\cite{SMILES} (SMILES) representation contains `.'. See Fig.~\ref{dataset_bars} for some other properties of these three datasets.

\subsubsection*{QM9}
The filtered QM9 dataset contains 133,015 molecules, each of which contains up to 9 heavy atoms of types \texttt{C}, \texttt{N}, \texttt{O} and \texttt{F}. Each molecule is paired with a reference conformation obtained by optimizing the molecular geometry with density functional theory (DFT) at the B3LYP/6-31G(2df,p) level of theory, {which implies that the reference conformations are obtained from the same environment.} We hold out separate 5,000 and 5,000 randomly selected molecules as validation and test sets, respectively.

\subsubsection*{COD}
We use the organic part of the COD dataset. We further filter out any molecule that contains more than 50 heavy atoms of types \texttt{B}, \texttt{C}, \texttt{N}, \texttt{O}, \texttt{F}, \texttt{Si}, \texttt{P}, \texttt{S}, \texttt{Cl}, \texttt{Ge}, \texttt{As}, \texttt{Se}, \texttt{Br}, \texttt{Te} and \texttt{I}. This results in 66,663 molecules, out of which we hold out separate 3,000 and 3,000 randomly selected ones respectively for validation and test purposes. Reference conformations are voluntarily contributed to the dataset and are often determined either experimentally or by DFT calculations.\cite{DFT} {Thus, the reference conformations are obtained from different environments.} 

\subsubsection*{CSD}
Similarly to COD, we remove any molecule that contains more than 50 heavy atoms, resulting in a total of 236,985 molecules. We hold out separate 3,000 and 3,000 randomly selected molecules for validation and test purposes respectively. This dataset contains organic and metal-organic crystallographic structures which have been observed experimentally.\cite{CSD} The atom types in this dataset are \texttt{S}, \texttt{N}, \texttt{P}, \texttt{Be}, \texttt{Tc}, \texttt{Xe}, \texttt{Br}, \texttt{Rh}, \texttt{Os}, \texttt{Zr}, \texttt{In}, \texttt{As}, \texttt{Mo}, \texttt{Dy}, \texttt{Nb}, \texttt{La}, \texttt{Te}, \texttt{Th}, \texttt{Ga}, \texttt{Tl}, \texttt{Y}, \texttt{Cr}, \texttt{F}, \texttt{Fe}, \texttt{Sb}, \texttt{Yb}, \texttt{Tb}, \texttt{Pu}, \texttt{Am}, \texttt{Re}, \texttt{Eu}, \texttt{Hg}, \texttt{Mn}, \texttt{Lu}, \texttt{Nd}, \texttt{Ce}, \texttt{Ge}, \texttt{Sc}, \texttt{Gd}, \texttt{Ca}, \texttt{Ti}, \texttt{Sn}, \texttt{Ir}, \texttt{Al}, \texttt{K}, \texttt{Tm}, \texttt{Ni}, \texttt{Er}, \texttt{Co}, \texttt{Bi}, \texttt{Pr}, \texttt{Rb}, \texttt{Sm}, \texttt{O}, \texttt{Pt}, \texttt{Hf}, \texttt{Se}, \texttt{Np}, \texttt{Cd}, \texttt{Pd}, \texttt{Pb}, \texttt{Ho}, \texttt{Ag}, \texttt{Mg}, \texttt{Zn}, \texttt{Ta}, \texttt{V}, \texttt{B}, \texttt{Ru}, \texttt{W}, \texttt{Cl}, \texttt{Au}, \texttt{U}, \texttt{Si}, \texttt{Li}, \texttt{C} and \texttt{I}. {The reference conformations are obtained from crystal structures.}

\subsection*{Models}

\subsubsection*{Baselines}

As a point of reference, we minimize a force field starting from a conformation created using ETKDG.\cite{ETKDG} We test both UFF and MMFF, and respectively call the resulting approaches {\bf ETKDG+UFF} and {\bf ETKDG+MMFF}. {The environment from which each conformation is obtained affects the force field calculations.} {To keep comparisons fair and to abstract away the effects of the environment,} we use the implementations in RDKit\cite{rdkit} with the default hyperparameters. {The default implementations have often been used in literature when comparing different conformation generation methods.\cite{defr1,defr2,defr3}}

\subsubsection*{Conditional Variational Graph Autoencoder}

We build one conditional variational graph autoencoder for each dataset. We use $d_h=50$ hidden units at each layer of neural message passing (Eq.~\ref{eq:mpnn}) in each of the three MPNNs corresponding to the prior, likelihood and posterior distributions. We use $d_f = 100$ in the two layer neural network that comes after the MPNN. As described earlier, we fix the variance of the output in the likelihood distribution to $1$. We use $L=3$ layers per network for QM9 and $L=5$ layers per network for COD and CSD. We chose these hyperparameter values by carrying out a grid-search and choosing the values that had the best performance on the validation set. The grid-search procedure and the performance of models with different hyperparameters are shown in the supplementary information.

\subsubsection*{Learning}

For all models, we use dropout~\cite{srivastava2014dropout} at each layer of the neural network that comes after the MPNN with a dropout rate of $0.2$ to regularize learning. We set the coefficient $\alpha$ in Eq.~\eqref{eq:final_loss} to $10^{-5}$. We train each model using Adam~\cite{kingma2014adam} with a fixed learning rate of $3 \times 10^{-4}$. All models were trained with a batch size of $20$ molecules on $1$ Nvidia GPU with $12$ GB of RAM.

\subsubsection*{Inference}

There are two modes of inference with the proposed approach. The first approach is to sample from a trained conditional variational graph autoencoder by first sampling from the prior distribution and taking the mean vectors from the likelihood distribution; we refer to this as {\bf CVGAE}. We can then use these samples further as initializations of MMFF minimization; we refer to this as {\bf CVGAE+MMFF}. The latter approach can be thought of as a trainable approach to initializing a conformation in place of DG or ETKDG.

\subsection*{Evaluation}

In principle, the quality of the sampled conformations should be evaluated based on their molecular energies, for instance by DFT, which is often more accurate than force field methods.\cite{lowenergyquant} However, the computational complexity of the DFT calculation is superlinear with respect to the number of electrons in a molecule, and so is often impractical.\cite{ratcliff2017challenges}. Instead, we follow prior work on conformation generation\cite{conformationgen_sota} and evaluate the baselines and proposed method using the RMSD (Eq.~\ref{eq:rmsd}) of the heavy atoms between a reference conformation and a predicted conformation which is fast and simple to calculate.

\section*{Results}

\begin{table*}[t]
	\label{table:results}
	\centering
    \caption{Number of successfully processed molecules in the test set (success per test set $\uparrow$), number of successfully generated conformations out of $100$ (success per molecule $\uparrow$), median of mean RMSD (mean $\downarrow$), median of standard deviation of RMSD (std. dev. $\downarrow$) and median of best RMSD (best $\downarrow$) per molecule on QM9, COD and CSD datasets. ETKDG stands for Distance Geometry with experimental torsion-angle preferences. UFF and MMFF are force field methods and stand for Universal Force Field and Molecular Mechanics Force Field respectively. CVGAE stands for Conditional Variational Graph Autoencoder. CVGAE + Force Field represents running the MMFF force field optimization initialized by CVGAE predictions.}
	\centering
    \begin{tabular}[b]{c|c|cc|c|c}
    \toprule
    & & \multicolumn{2}{c|}{ETKDG + Force Field} & \multicolumn{1}{c|}{CVGAE} & \multicolumn{1}{c}{CVGAE + Force Field} \\
    Dataset & & UFF & MMFF &  & MMFF  \\
    \midrule
    QM9 & success per test set  & 96.440$\%$ & 96.440$\%$ & 100$\%$ & 99.760$\%$ \\
     & success per molecule  & 98.725$\%$ & 98.725$\%$ & 100$\%$ & 98.684$\%$ \\
     & mean & 0.425 &	0.415 &	0.390 &	0.367 \\
     & std. dev. & 0.176 &	0.189 &	0.017 &	0.074 \\
     & best & 0.126 & 0.092 & 0.325 & 0.115 \\
     \midrule
    COD & success per test set  & 99.133$\%$ & 99.133$\%$ & 100$\%$ & 95.367$\%$ \\
     & success per molecule  & 99.627$\%$ & 99.627$\%$ & 100$\%$ & 99.071$\%$ \\
     & mean  & 1.389 & 1.358 & 1.331 & 1.656 \\
     & std. dev.  & 0.407 & 0.415 & 0.099 & 0.425 \\
     & best & 0.429	& 0.393 & 1.206	 & 0.635 \\
     \midrule
    CSD & success per test set  & 97.400$\%$ & 97.400$\%$ & 100$\%$ & 99.467$\%$ \\
     & success per molecule  & 99.130$\%$ & 99.130$\%$ & 100$\%$ & 97.967$\%$ \\
     & mean  & 1.537 & 1.488 & 1.506 & 1.833 \\
     & std. dev.  & 0.421 & 0.418 & 0.115 & 0.434 \\
     & best  & 0.508 & 0.478 & 1.343 & 0.784 \\
    \bottomrule
    \end{tabular}
    \label{tab:results_rmsd}
\end{table*}    

When evaluating each method, we first sample $100$ conformations per molecule for each method in the test set. We can make several observations from Table \ref{tab:results_rmsd}. First, compared to other methods, our proposed CVGAE always succeeds at generating the specified number of conformations for any of the molecules in the test set. UFF and MMFF fail to generate conformations for some molecules, as they do not support handling every element but the pre-defined sets of elements. Since all other evaluated approaches were unsuccessful at generating at least one conformation for a very small number of test molecules, we report results for the molecules for which all evaluated methods generated at least one conformation. We report the \textit{median} of the mean of the RMSD, the \textit{median} of the standard deviation of the RMSD and the \textit{median} of the best (lowest) RMSD among all generated conformations for each test molecule. Across all three datasets, every evaluated method achieves roughly the same median of the mean RMSD. More importantly, the standard deviation of the RMSD achieved by CVGAE is \textit{significantly} lower than that achieved by ETKDG + Force Field. After the initial generation stage, conformations are usually further evaluated and optimized by running the computationally expensive DFT optimization. Reducing the standard deviation can lower the number of conformations on which DFT optimization has to be run in order to achieve a valid conformation. On the other hand, the best RMSD achieved by ETKDG + UFF/MMFF methods is lower than that achieved by CVGAE. Using MMFF initialized by CVGAE (CVGAE + MMFF) instead of ETKDG (ETKDG + MMFF) improves the mean results on the QM9 dataset for CVGAE, and yields a lower standard deviation and similar best RMSD compared to ETKDG + MMFF. Unfortunately, CVGAE + MMFF worsens the results achieved by CVGAE alone on the COD and CSD datasets. We additionally evaluate single point DFT energy for the subset of $1000$ molecules in the QM9 test set for all $100$ generated conformations. We find that all three methods ETKDG + MMFF, CVGAE and CVGAE + MMFF achieve similar median energy values of $-411.52$, $-410.87$ and $-411.50$ respectively.  The energy was calculated using GAMESS software \cite{Schmidt1993GAMESS} with default parameters.

\begin{table}[t]
    \caption{Conformation Diversity. Mean and std. dev. represents the corresponding mean and standard deviation of pairwise RMSD between at most $100$ generated conformations per molecule.}
	\label{table:results2}
	\centering
    \begin{tabular}[b]{c|c|c|c|c}
    \toprule
    Dataset & & ETKDG + MMFF & CVGAE & CVGAE + MMFF  \\
    \midrule
    QM9 & mean &  0.400 & 0.106 & 0.238 \\
     & std. dev. &  0.254 & 0.061 & 0.209 \\
     \midrule
    COD & mean  &  1.148 & 0.239 & 1.619 \\
     & std. dev.  &  0.699 & 0.181 & 0.537 \\
     \midrule
    CSD & mean  & 1.244 & 0.567 & 1.665 \\
     & std. dev.  &  0.733 & 0.339 & 0.177 \\
     \bottomrule
    \end{tabular}
    \label{tab:rmsd_diversity}
\end{table}    

We also report the diversity of conformations generated by all evaluated methods in Table \ref{tab:rmsd_diversity}. Diversity is measured by calculating the mean and standard deviation of the pairwise RMSD between each pair of generated conformations per molecule. Overall, we can see that despite having a smaller median of standard deviation of RMSD between generated conformations and reference conformations, CVGAE does not collapse to generating extremely similar conformations. Although, CVGAE generates relatively less diverse samples compared to ETKDG + MMFF baseline on all datasets. The conformations of molecules generated by CVGAE + MMFF are less diverse on the QM9 dataset and more diverse on COD/CSD datasets compared to ETKDG + MMFF baseline.
\begin{figure}[t!]
    \begin{minipage}{0.48\textwidth}
    \centering
        \includegraphics[width=\textwidth]{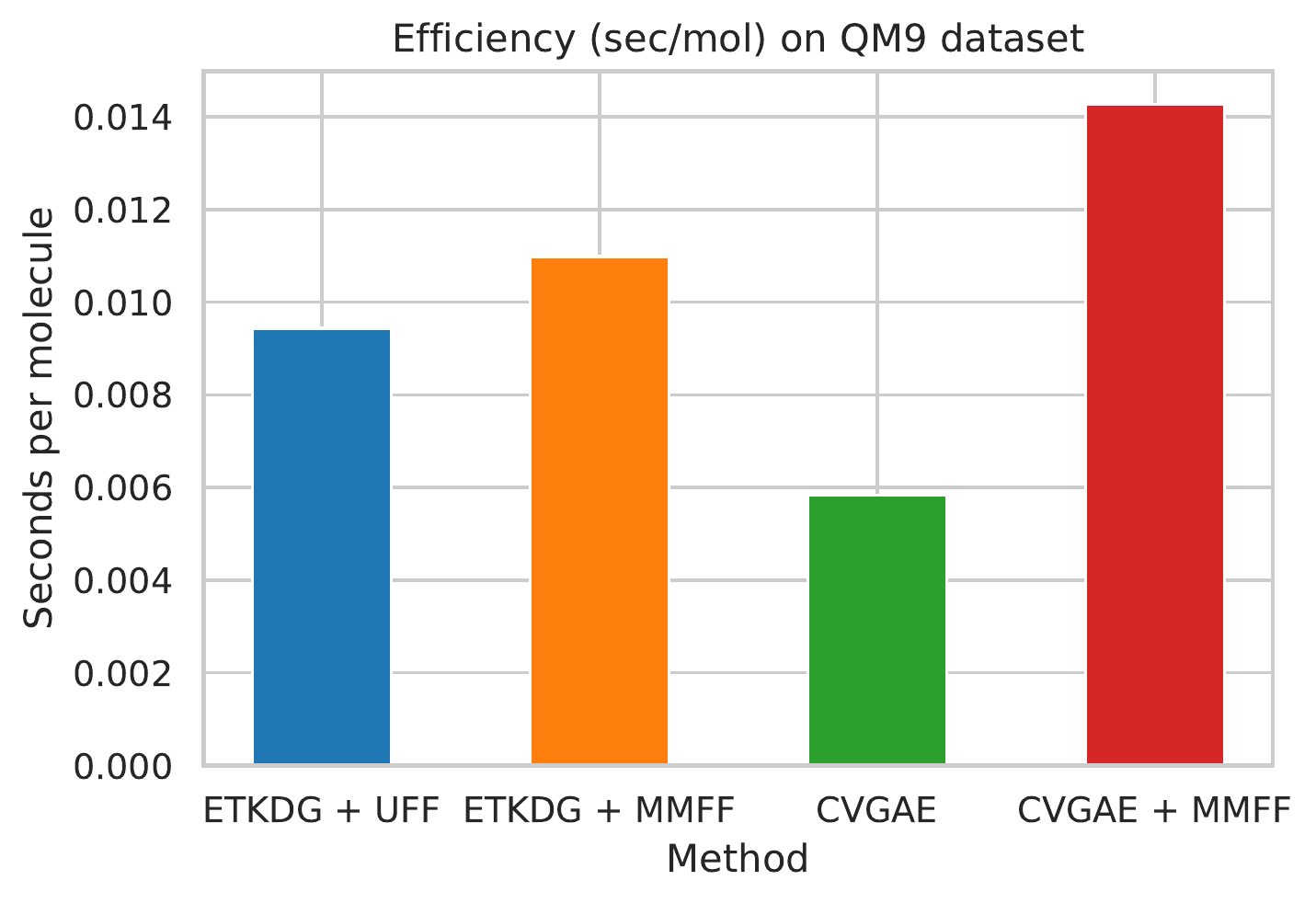}
        (a) QM9 dataset
    \end{minipage}
    \begin{minipage}{0.48\textwidth}
    \centering
        \includegraphics[width=\textwidth]{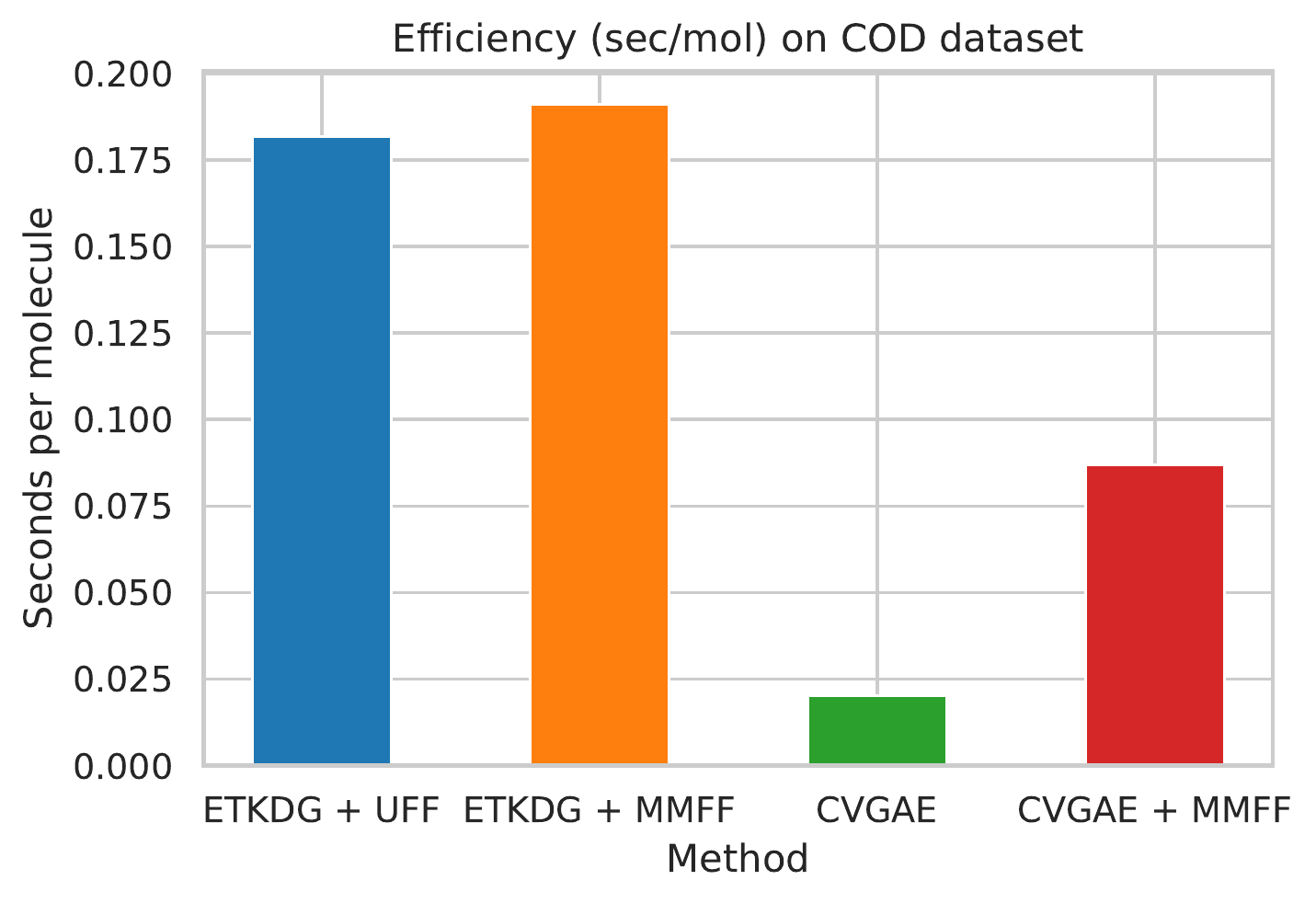}
        (b) COD dataset
    \end{minipage}
    \caption{Computational efficiency of various approaches on QM9 and COD datasets}
    \label{dataset_speed}
\end{figure}

The computational efficiency of each of the evaluated approaches on the QM9 and COD datasets is shown in Figure \ref{dataset_speed}. For consistency, we generated one conformation for one molecule at a time using each of the evaluated methods on an Intel(R) Xeon(R) E5-2650  v4 CPU. On the QM9 dataset, CVGAE is $2\times$ more efficient than ETKDG + UFF/MMFF, while CVGAE + MMFF is slightly slower than ETKDG + UFF/MMFF. On the COD dataset, which contains a larger number of atoms per molecule, CVGAE is almost $10\times$ as fast as ETKDG + UFF/MMFF, while CVGAE + MMFF is about $2\times$ as fast as ETKDG + UFF/MMFF. This shows that CVGAE scales much better than the baseline ETKDG + UFF/MMFF methods as the size of the molecule grows.

\begin{figure}
    \centering
    \begin{subfigure}{0.45\textwidth}
        \includegraphics[width=\textwidth]{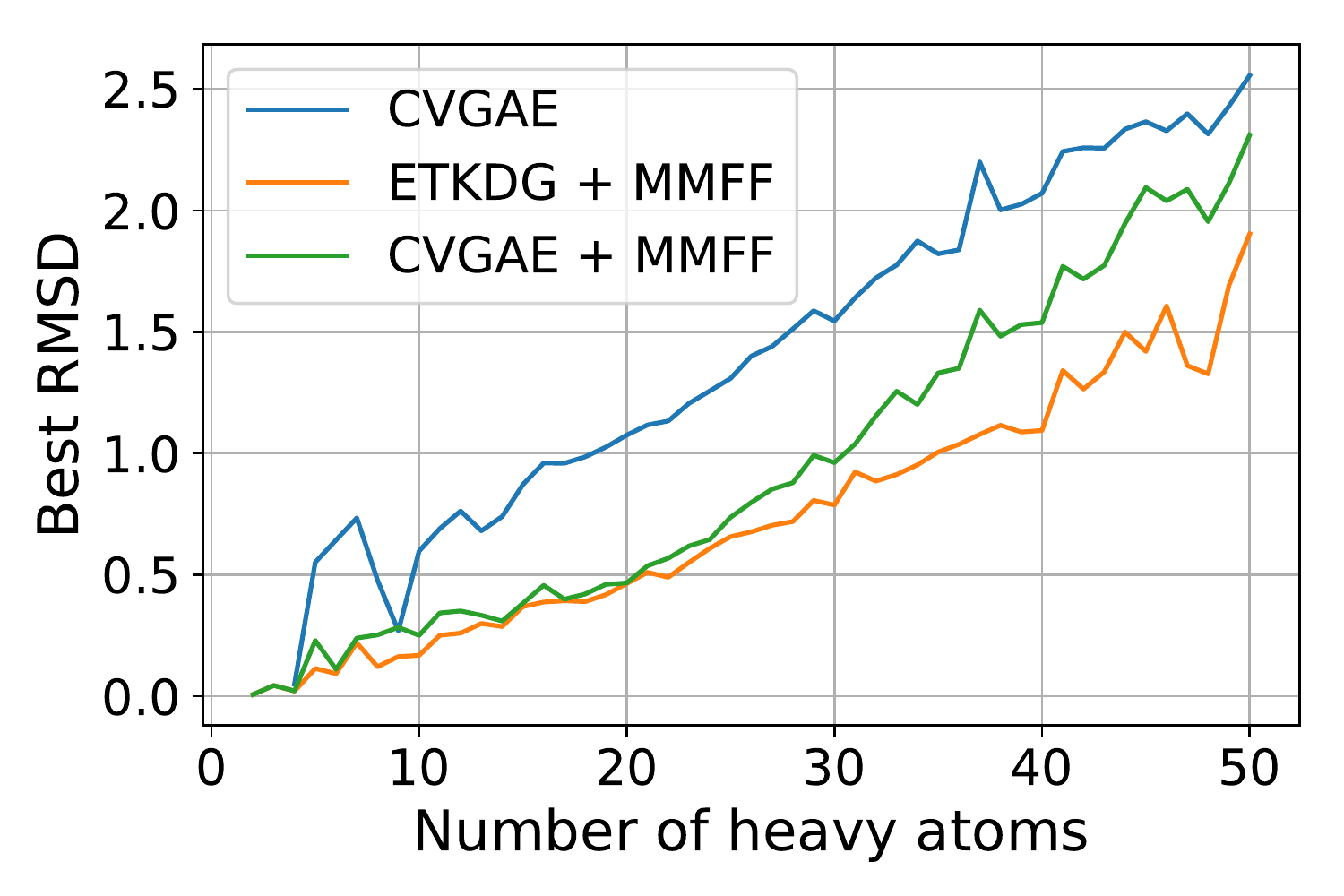}
        \caption{Mean of best RMSD}
        \label{c_best_rmsd_v_ha}
    \end{subfigure}
    \begin{subfigure}{0.45\textwidth}
        \includegraphics[width=\textwidth]{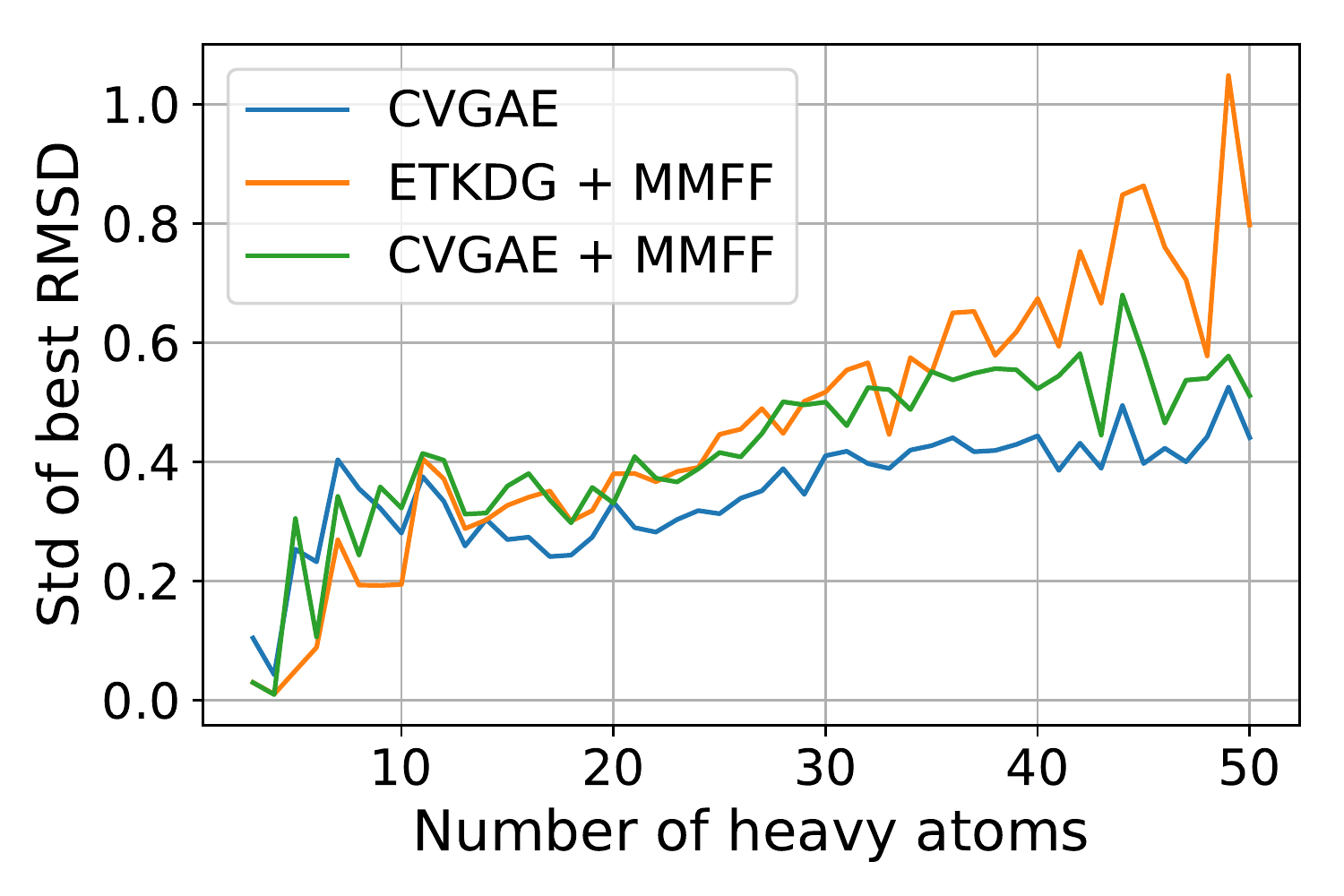}
        \caption{St. dev. of best RMSD}
        \label{c_std_best_rmsd_v_ha_trunc}
    \end{subfigure}
    \begin{subfigure}{0.45\textwidth}
        \includegraphics[width=\textwidth]{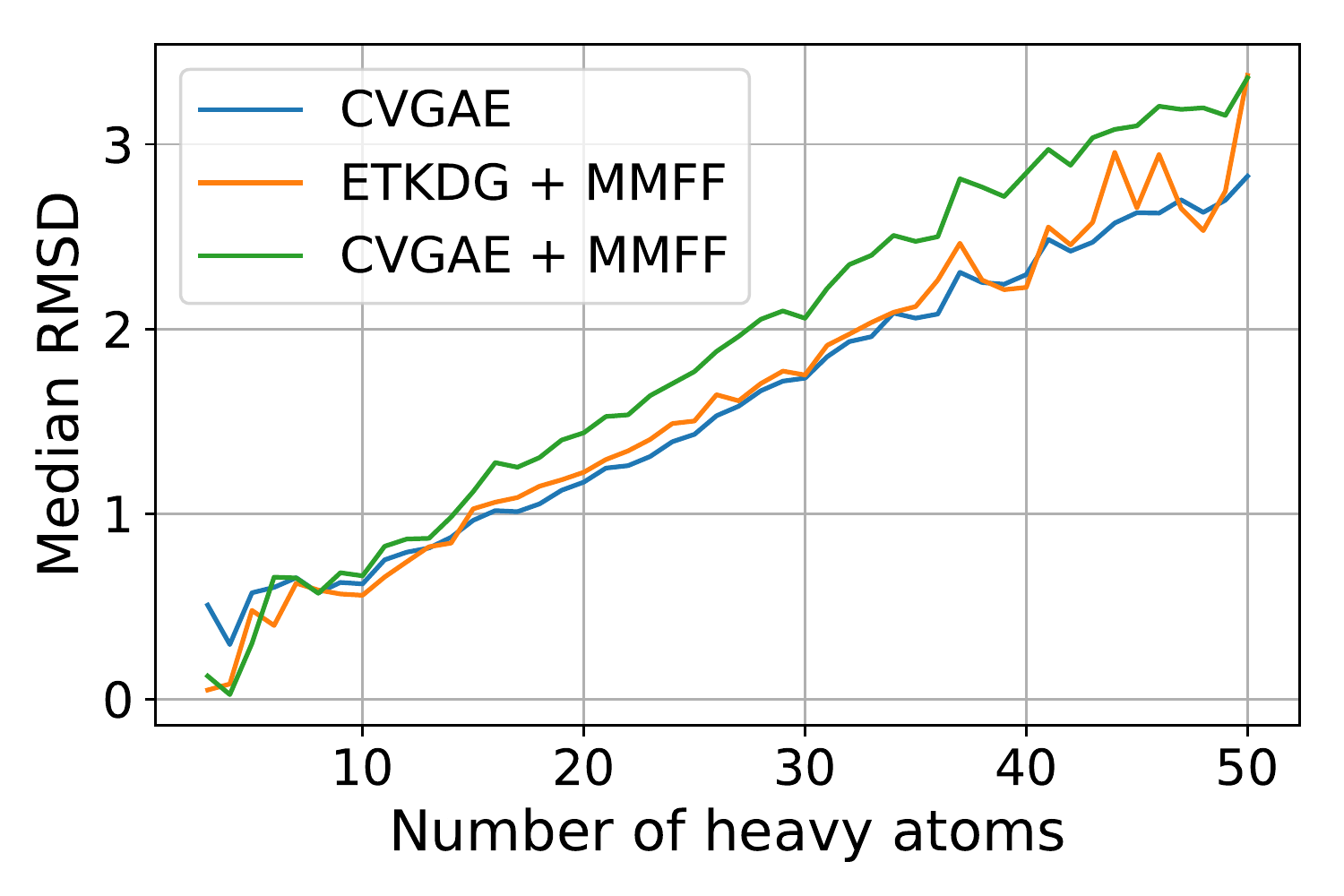}
        \caption{Mean of median RMSD}
        \label{c_median_rmsd_v_ha_trunc}
    \end{subfigure}
    \begin{subfigure}{0.45\textwidth}
        \includegraphics[width=\textwidth]{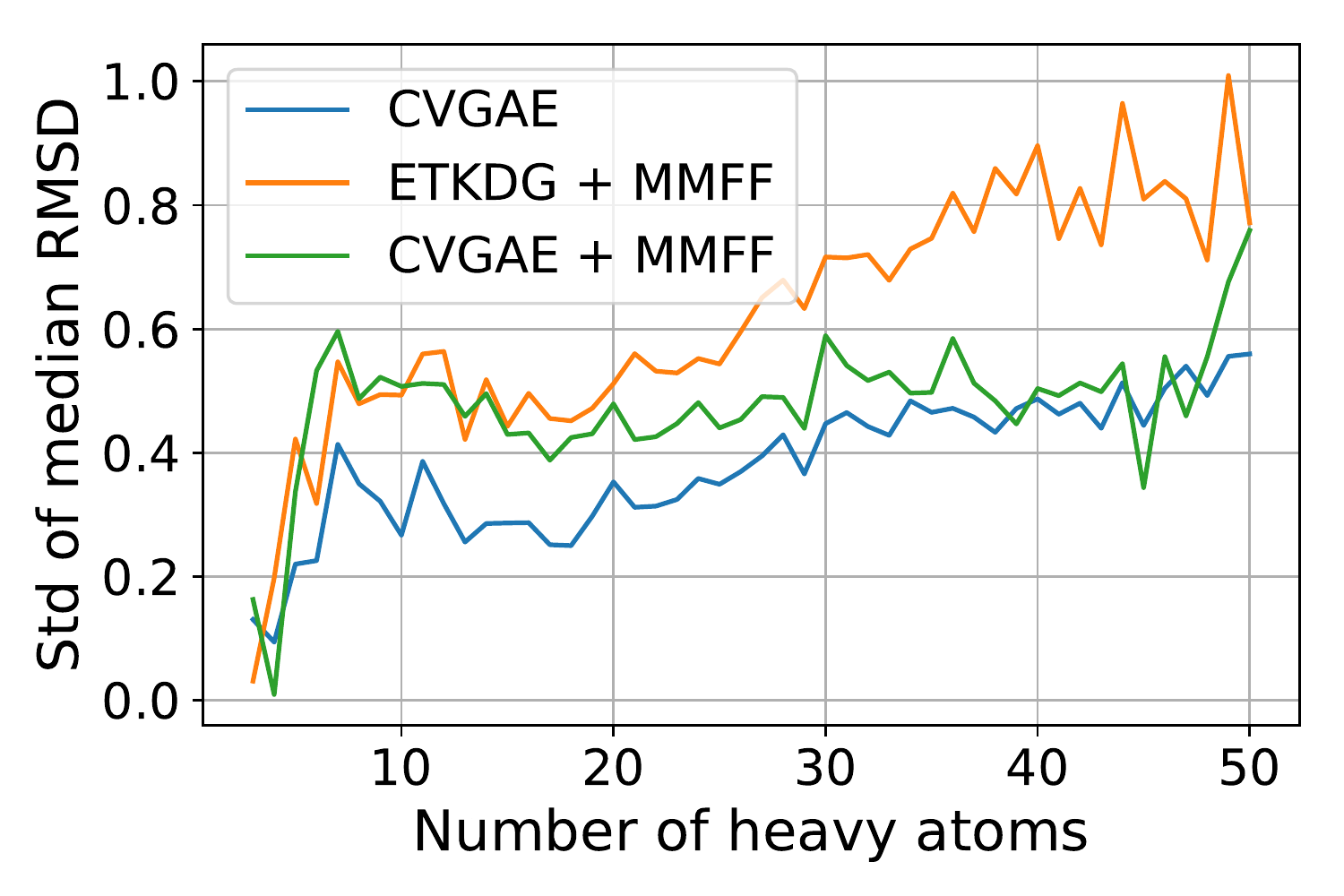}
        \caption{St. dev. of median RMSD}
        \label{c_std_median_rmsd_v_ha_trunc}
    \end{subfigure}
    \begin{subfigure}{0.45\textwidth}
        \includegraphics[width=\textwidth]{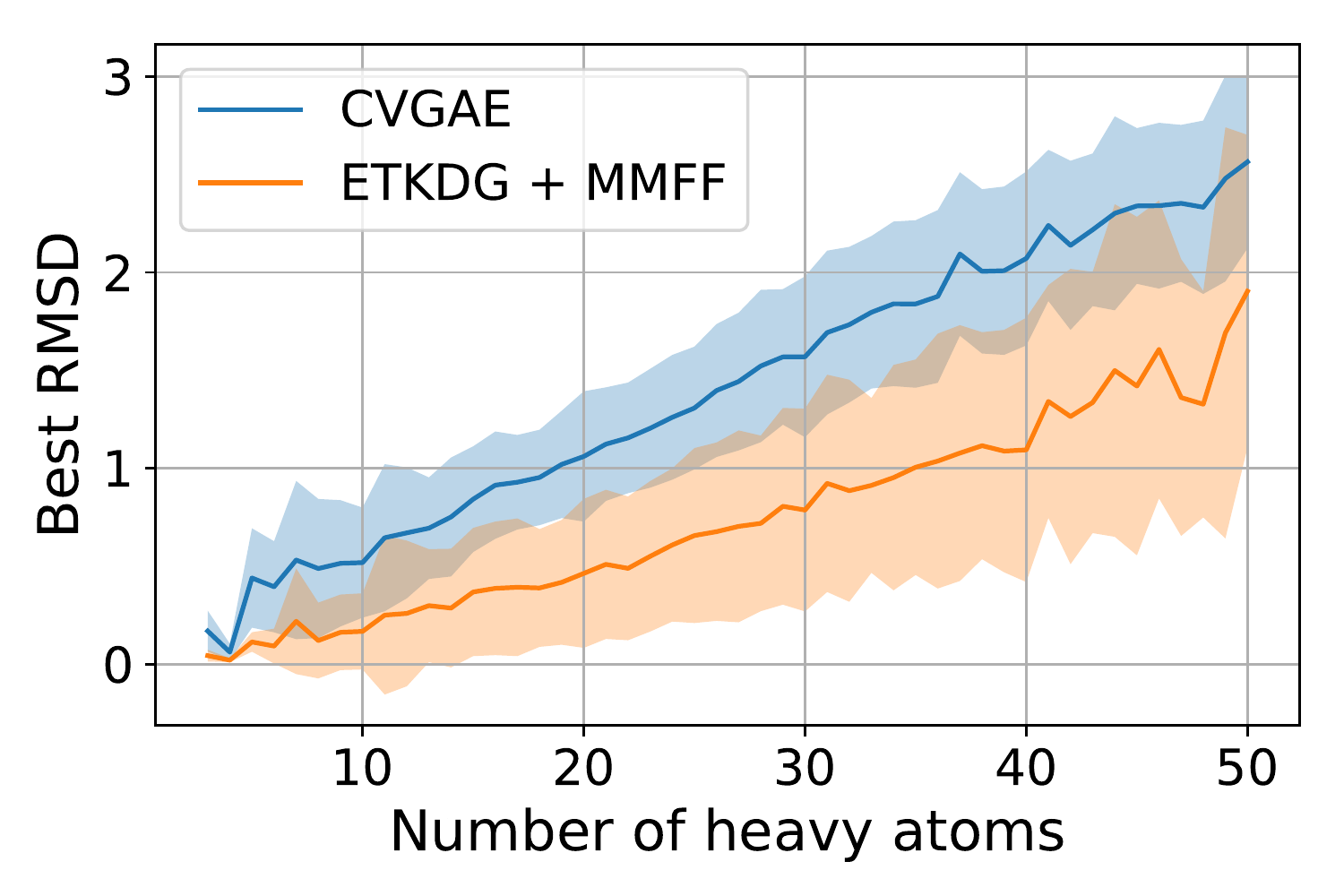}
        \caption{Best RMSD with uncertainty bounds}
        \label{c_best_rmsd_error_v_ha_trunc}
    \end{subfigure}
    \begin{subfigure}{0.45\textwidth}
        \includegraphics[width=\textwidth]{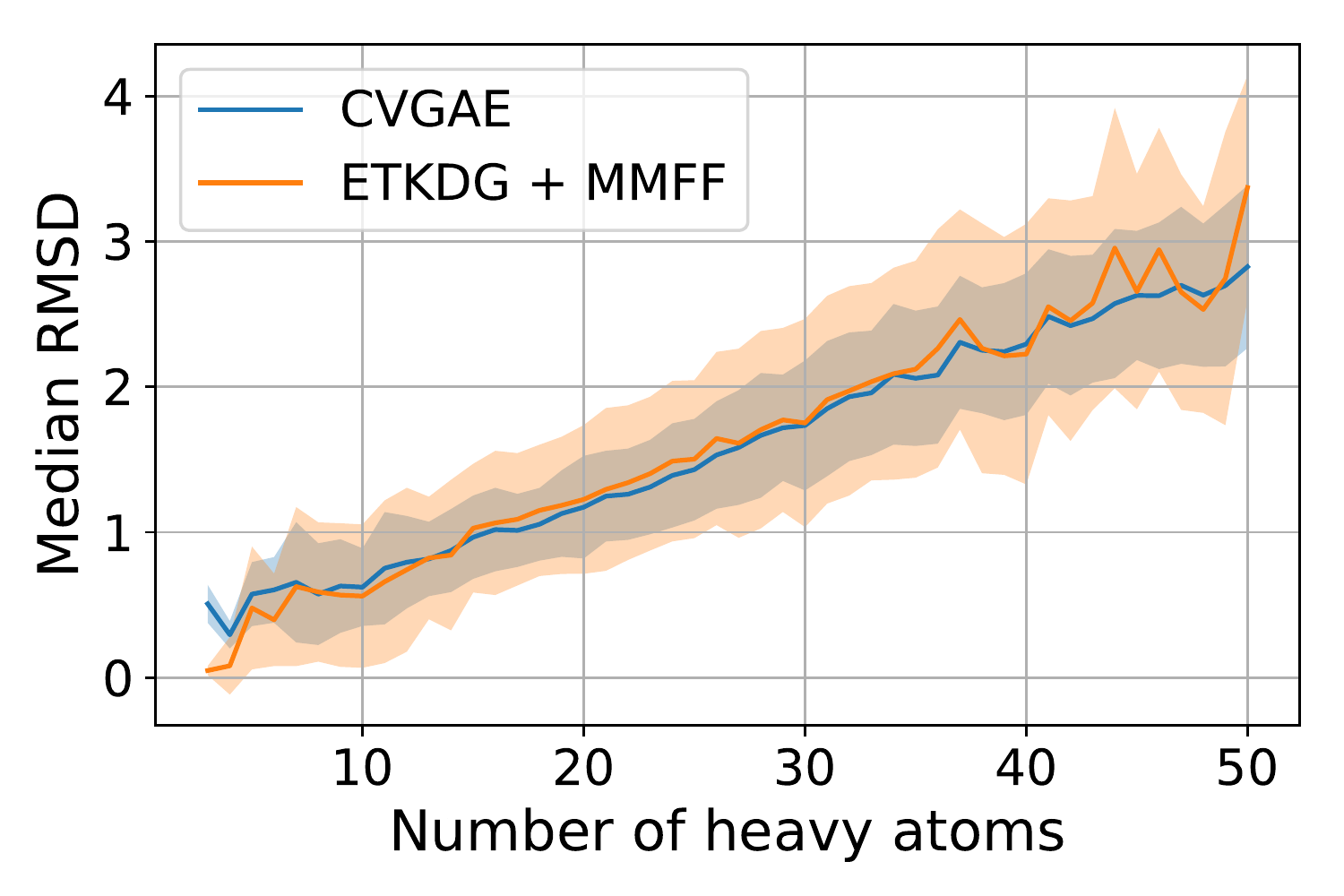}
        \caption{Median RMSD with uncertainty bounds}
        \label{c_median_rmsd_error_v_ha_trunc}
    \end{subfigure}
    \caption{This figure shows the means and standard deviations of the best and median RMSDs on the union of COD and CSD datasets as a function of number of heavy atoms. The molecules were grouped by number of heavy atoms, and the mean and standard deviation of the median and best RMSDs were calculated for each group to obtain these plots. Groups at the left hand side of the graph with less than 1\% of the mean number of molecules per group were omitted.}
    \label{c_rmsd_v_ha_trunc}
\end{figure}

\begin{figure}
    \centering
    \begin{subfigure}{0.45\textwidth}
        \includegraphics[width=\textwidth]{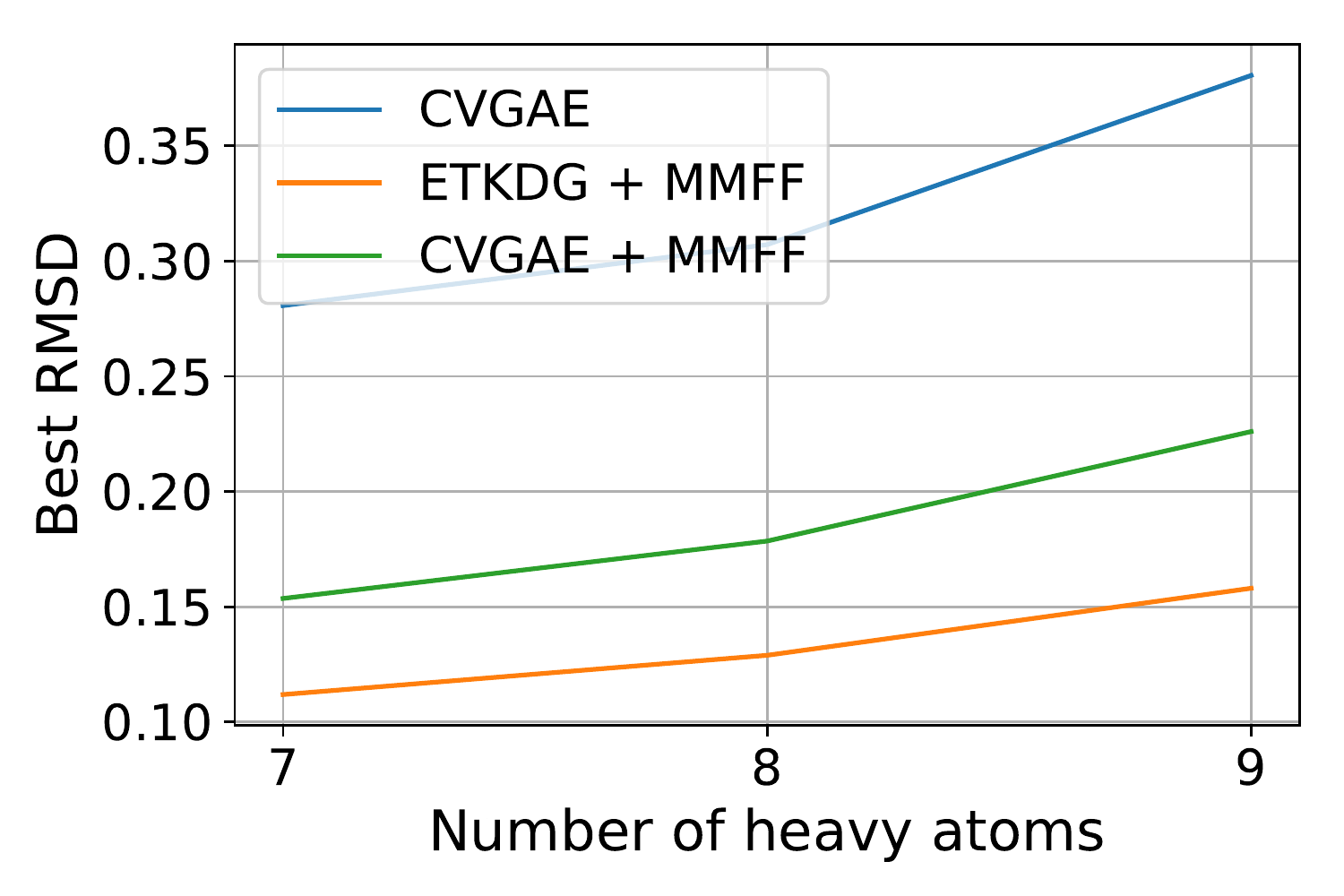}
        \caption{Mean of best RMSD}
        \label{q_best_rmsd_v_ha_trunc}
    \end{subfigure}
    \begin{subfigure}{0.45\textwidth}
        \includegraphics[width=\textwidth]{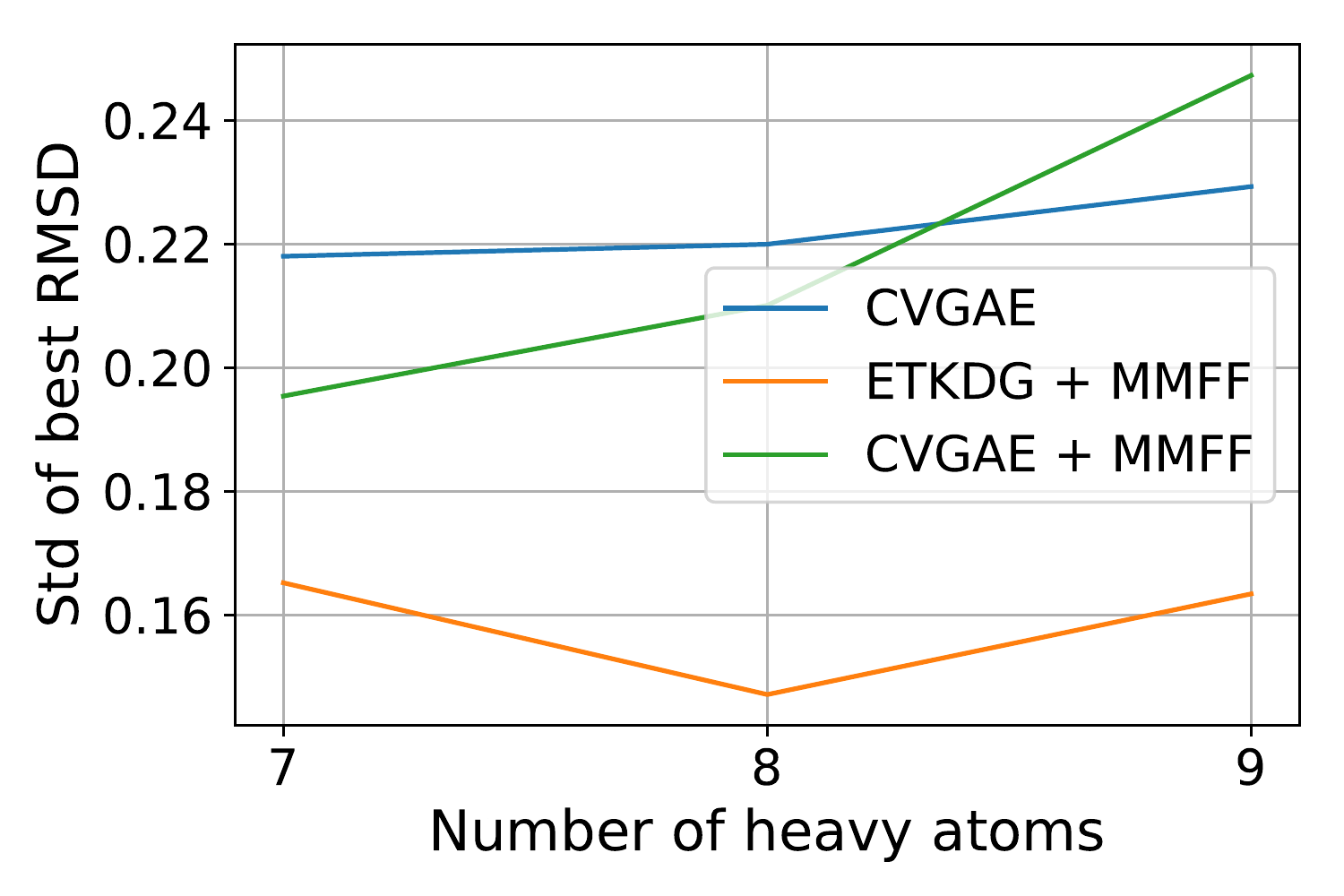}
        \caption{St. dev. of best RMSD}
        \label{q_std_best_rmsd_v_ha_trunc}
    \end{subfigure}
    \begin{subfigure}{0.45\textwidth}
        \includegraphics[width=\textwidth]{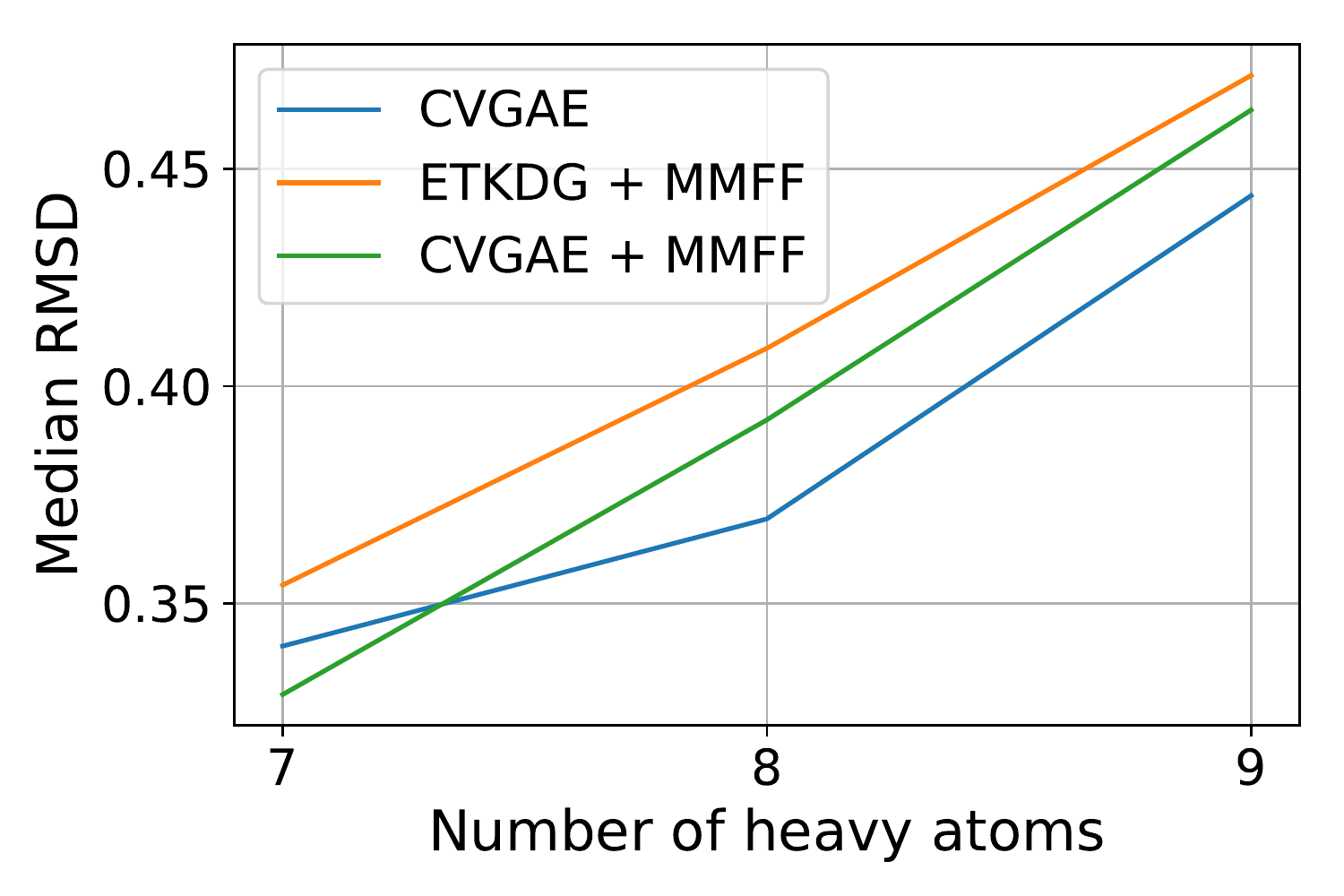}
        \caption{Mean of median RMSD}
        \label{q_median_rmsd_v_ha_trunc}
    \end{subfigure}
    \begin{subfigure}{0.45\textwidth}
        \includegraphics[width=\textwidth]{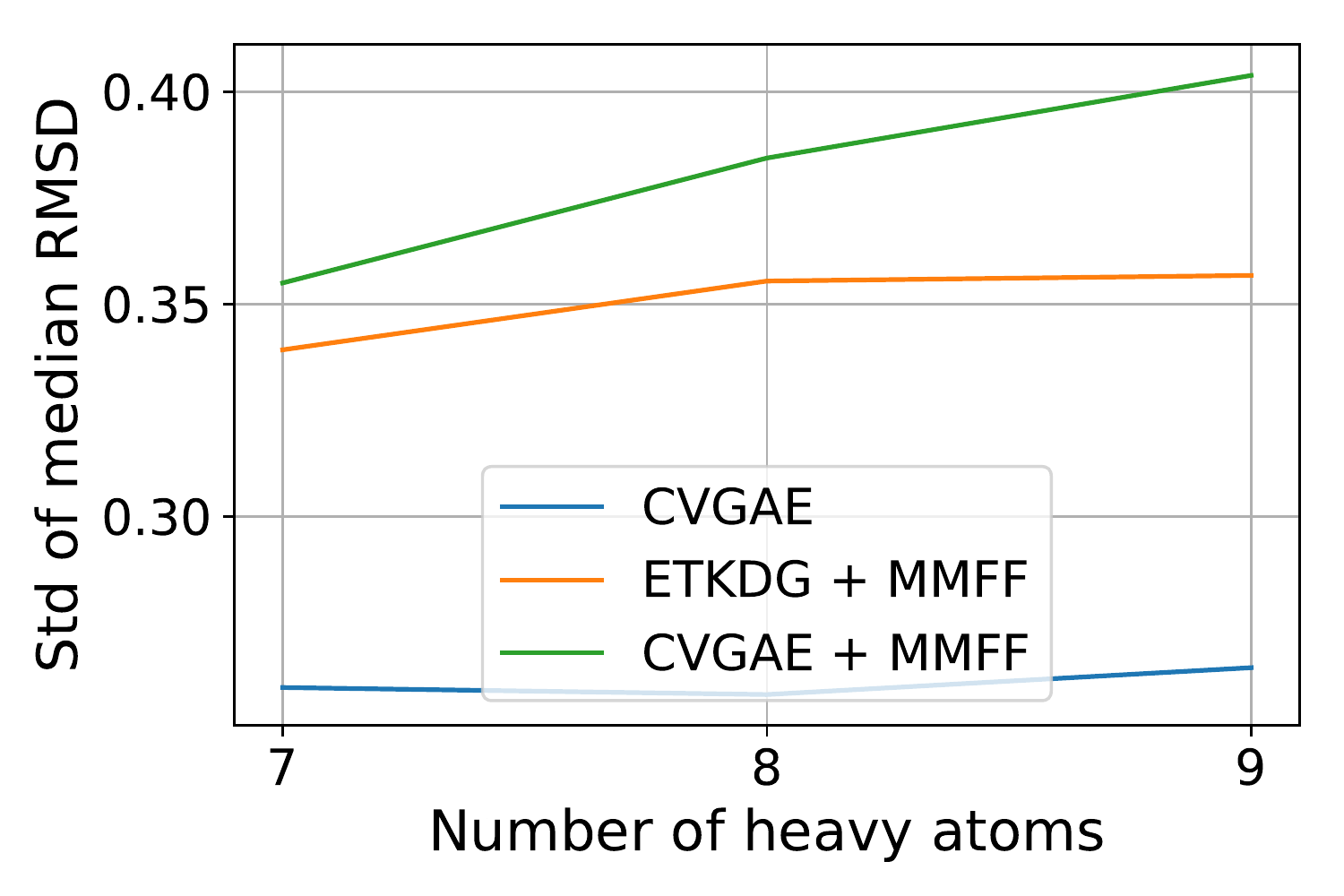}
        \caption{St. dev. of median RMSD}
        \label{q_std_median_rmsd_v_ha_trunc}
    \end{subfigure}
    \begin{subfigure}{0.45\textwidth}
        \includegraphics[width=\textwidth]{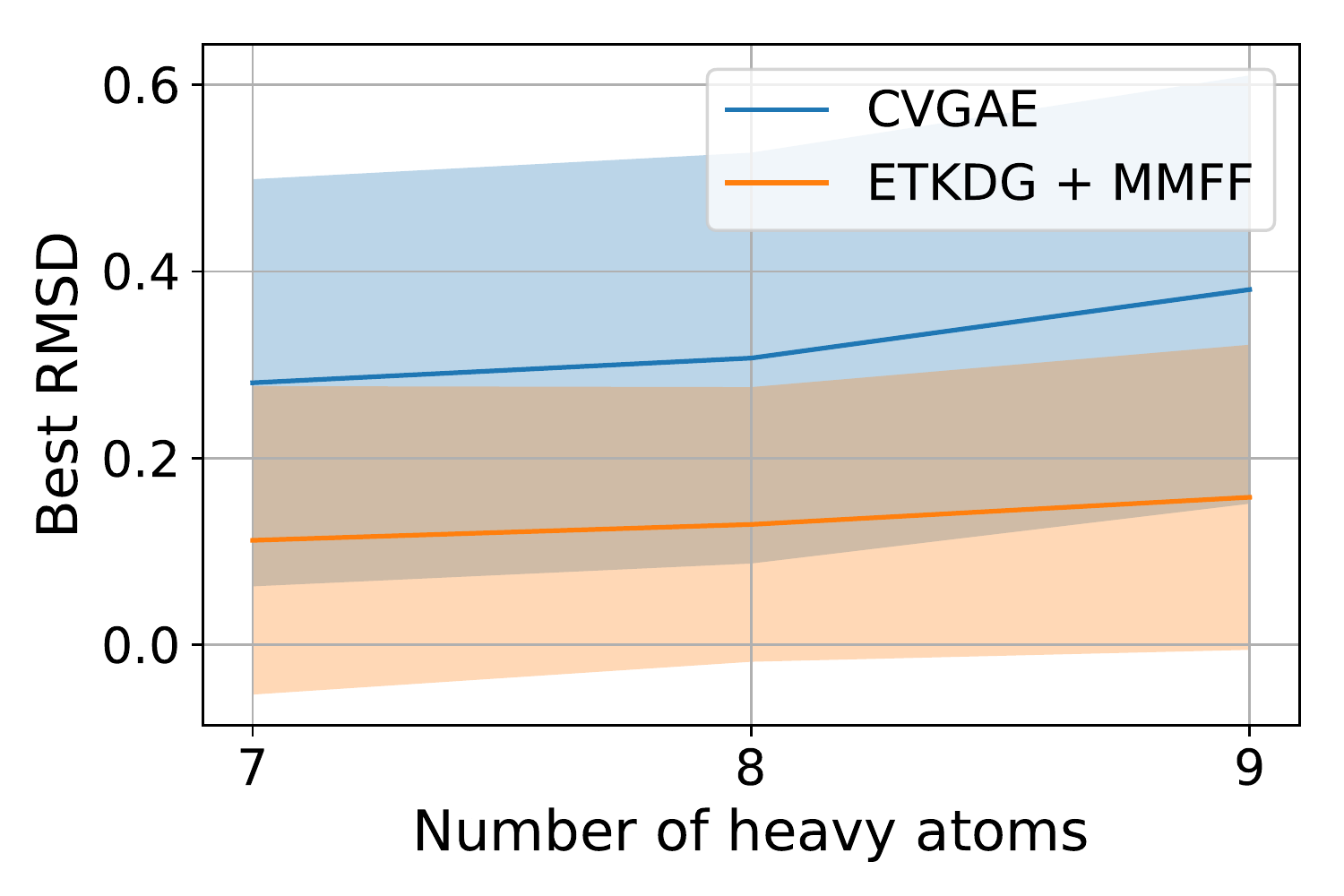}
        \caption{Best RMSD with uncertainty bounds}
        \label{q_best_rmsd_error_v_ha_trunc}
    \end{subfigure}
    \begin{subfigure}{0.45\textwidth}
        \includegraphics[width=\textwidth]{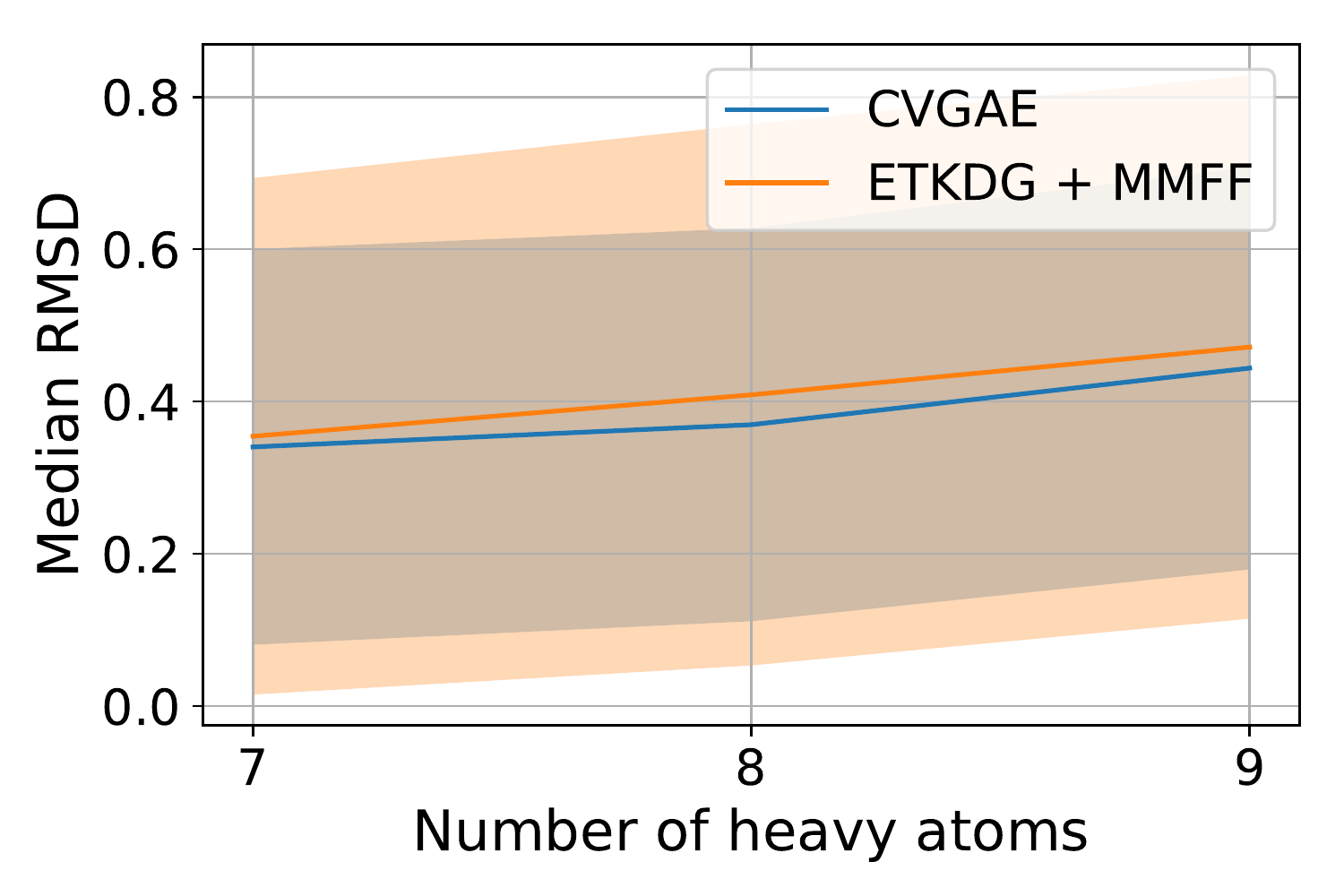}
        \caption{Median RMSD with uncertainty bounds}
        \label{q_median_rmsd_error_v_ha_trunc}
    \end{subfigure}
    \caption{This figure shows the means and standard deviations of the best and median RMSDs on the QM9 dataset as a function of number of heavy atoms. The molecules were grouped by number of heavy atoms, and the mean and standard deviation of the median and best RMSDs were calculated for each group to obtain these plots. Groups at the left hand side of the graph with less than 1\% of the mean number of molecules per group were omitted.}
    \label{q_rmsd_v_ha_trunc}
\end{figure}

Figures \ref{c_rmsd_v_ha_trunc} and \ref{q_rmsd_v_ha_trunc} visualize the median, standard deviation and best RMSD results as a function of the number of heavy atoms in a molecule on the QM9 and COD/CSD datasets respectively. For all approaches, we can see that the best and median RMSD both increase with the number of heavy atoms. The standard deviation of the median RMSD for CVGAE and CVGAE + MMFF is lower than that for ETKDG + MMFF across molecules of almost all sizes. The standard deviation of the best RMSD is slightly higher for CVGAE and CVGAE + MMFF than for ETKDG + MMFF on molecules with at most 12 atoms, but is lower for larger molecules, particularly for CVGAE. Overall, CVGAE yields a lower or similar median RMSD compared to ETKDG + MMFF across molecules of all sizes and a lower standard deviation, whereas ETKDG + MMFF provides a lower best RMSD particularly for larger molecules observed in the COD/CSD datasets.

\begin{figure}[htbp]
    \begin{subfigure}{0.45\textwidth}
        \includegraphics[width=\textwidth,height=0.22\textheight]{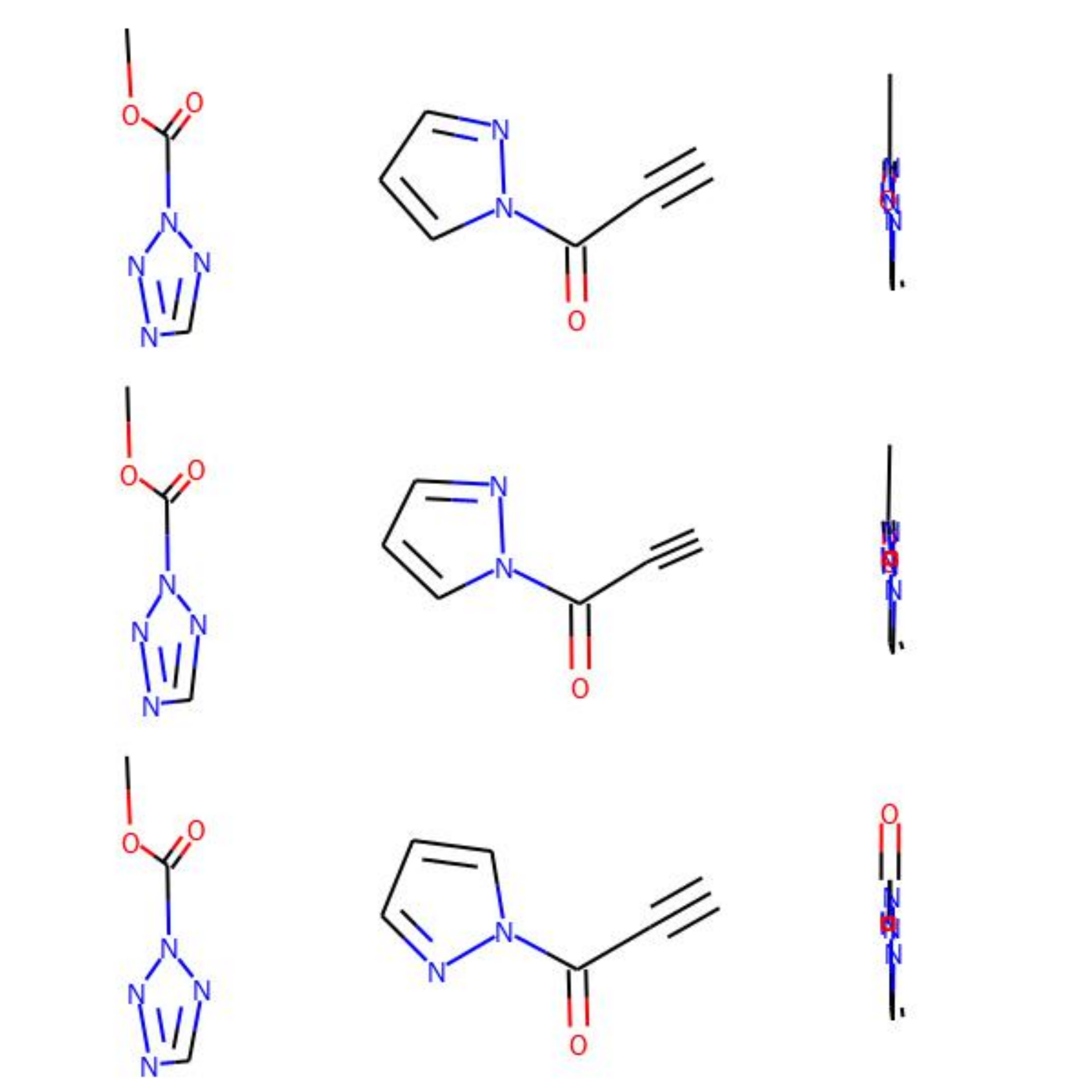}
        \caption{QM9 greatest difference in favour of neural network predictions}
        \label{QM9 best nnvff}
    \end{subfigure}
    \rulesep
    \begin{subfigure}{0.45\textwidth}
        \includegraphics[width=\textwidth,height=0.22\textheight]{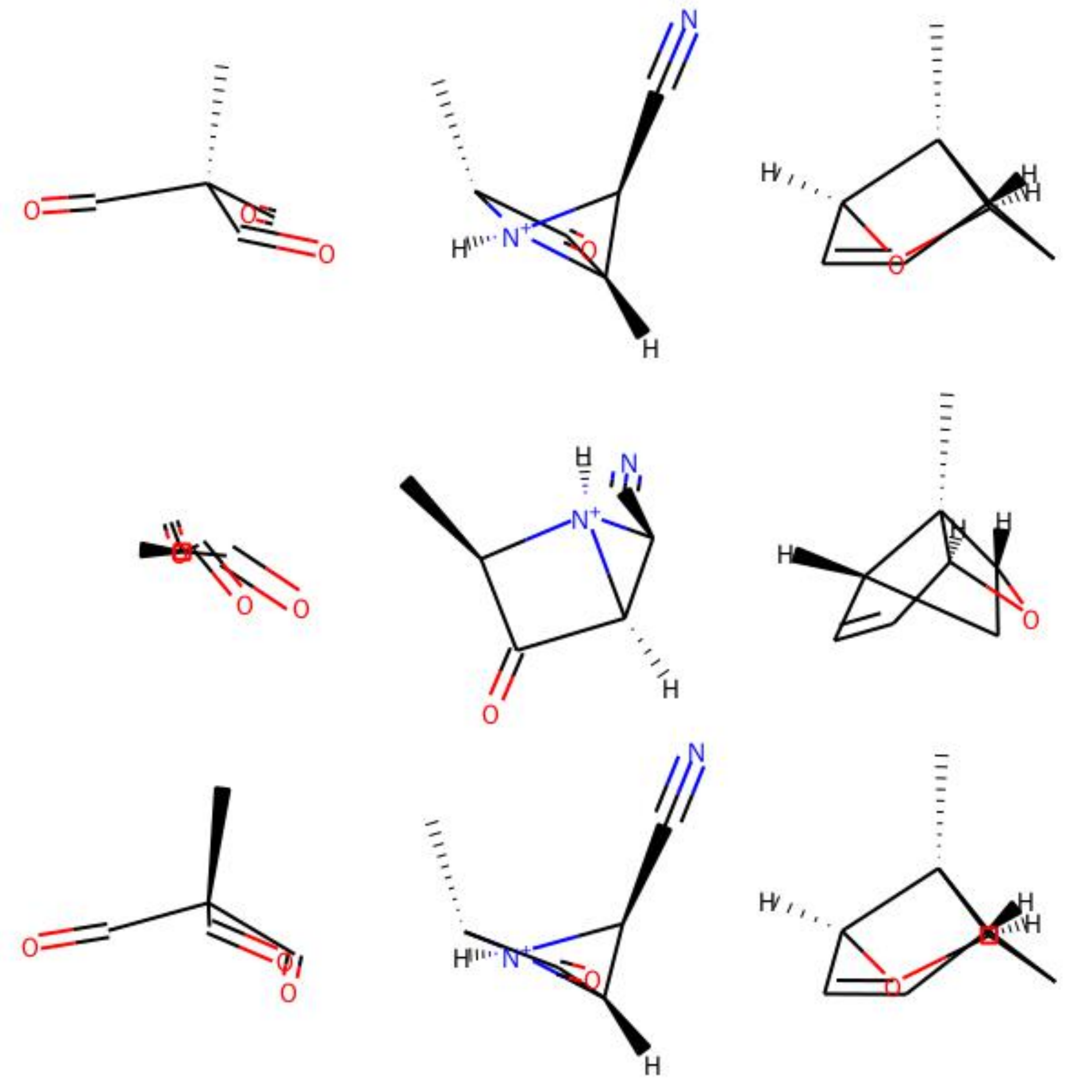}
        \caption{QM9 greatest difference in favour of ETKDG + MMFF predictions}
        \label{QM9 worst nnvff}
    \end{subfigure}
    \hrule
    \begin{subfigure}{0.45\textwidth}
        \includegraphics[width=\textwidth,height=0.22\textheight]{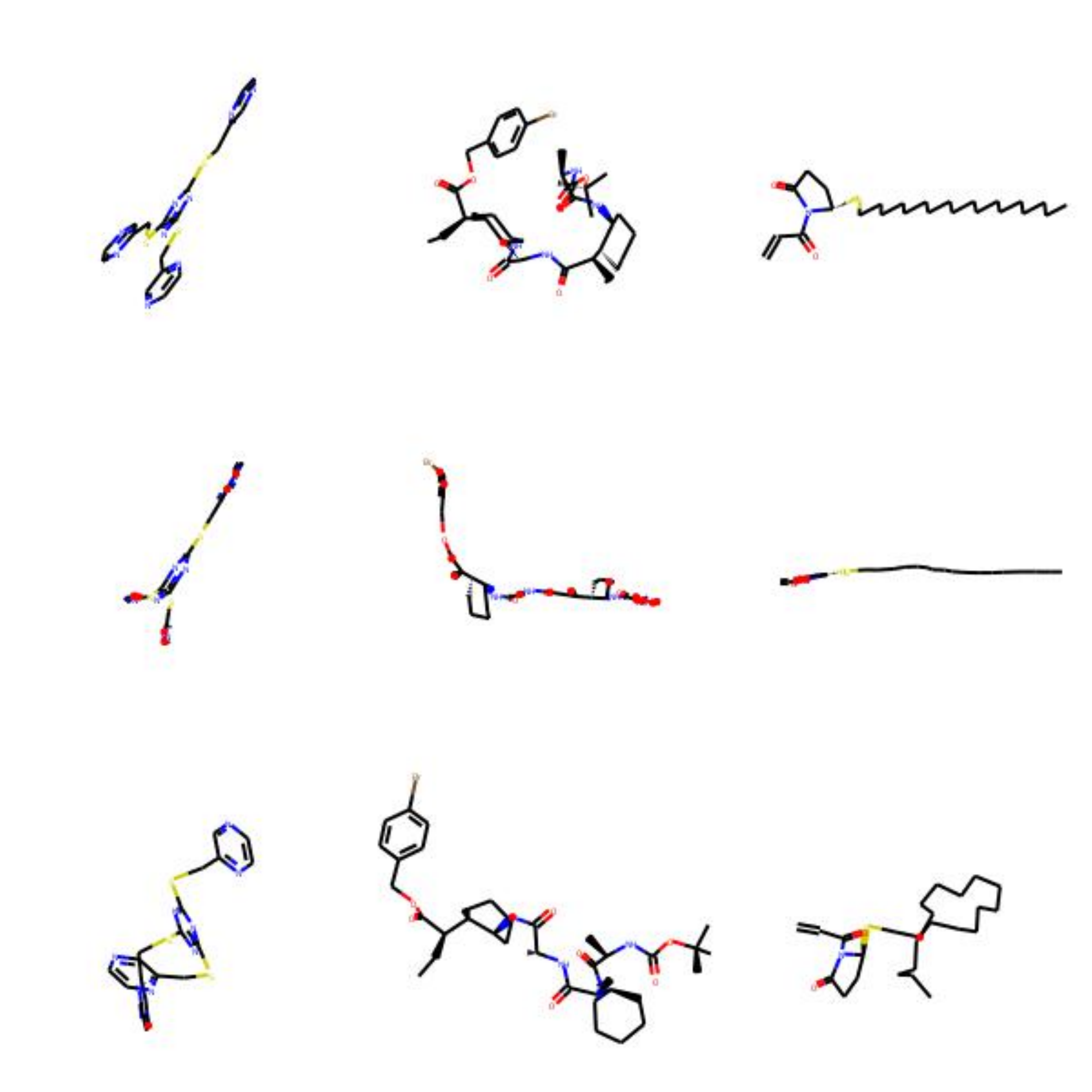}
        \caption{COD greatest difference in favour of neural network predictions}
        \label{COD best nnvff}
    \end{subfigure}
    \rulesep
    \begin{subfigure}{0.45\textwidth}
        \includegraphics[width=\textwidth,height=0.22\textheight]{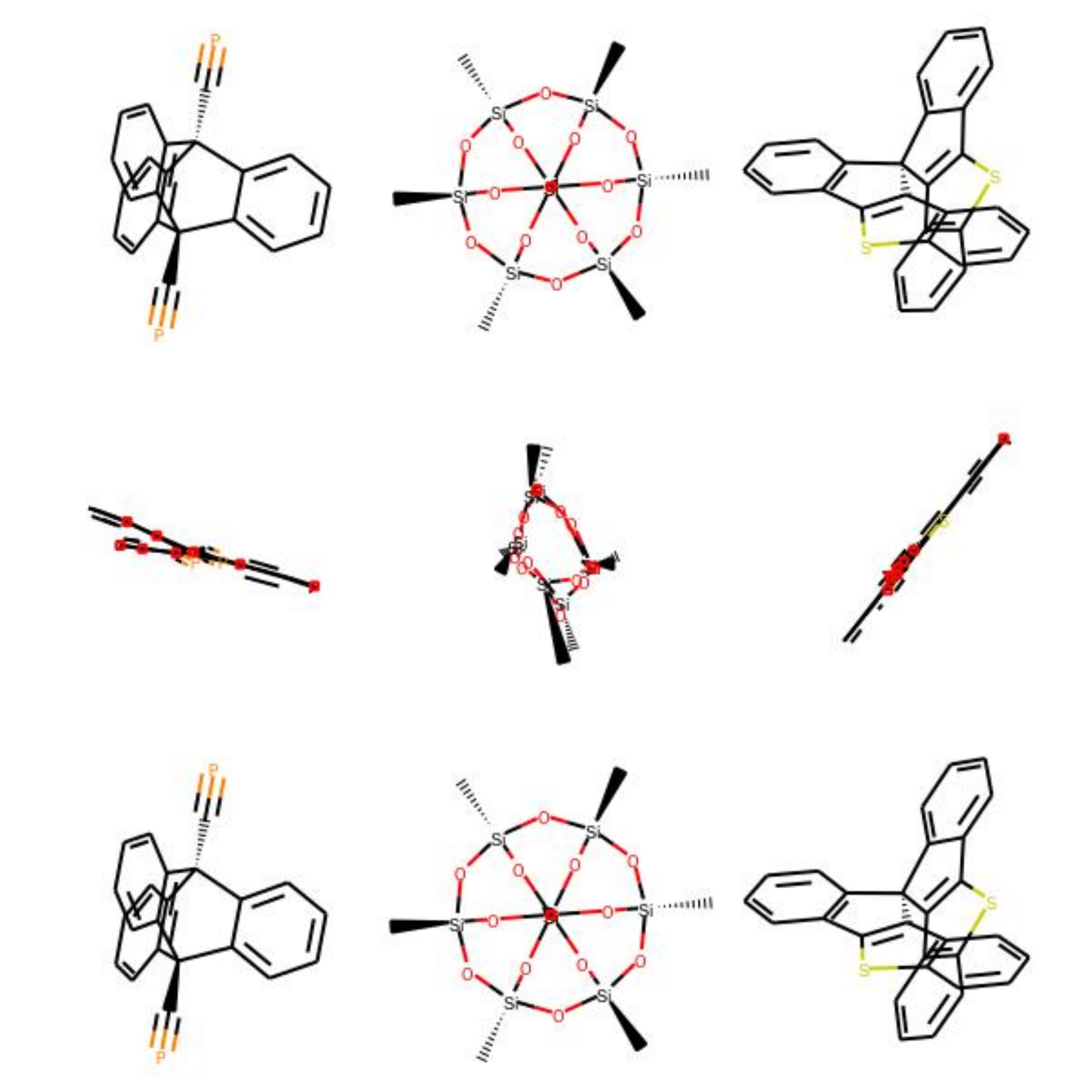}
        \caption{COD greatest difference in favour of ETKDG + MMFF predictions}
        \label{COD worst nnvff}
    \end{subfigure}
    \hrule
    \begin{subfigure}{0.45\textwidth}
        \includegraphics[width=\textwidth,height=0.22\textheight]{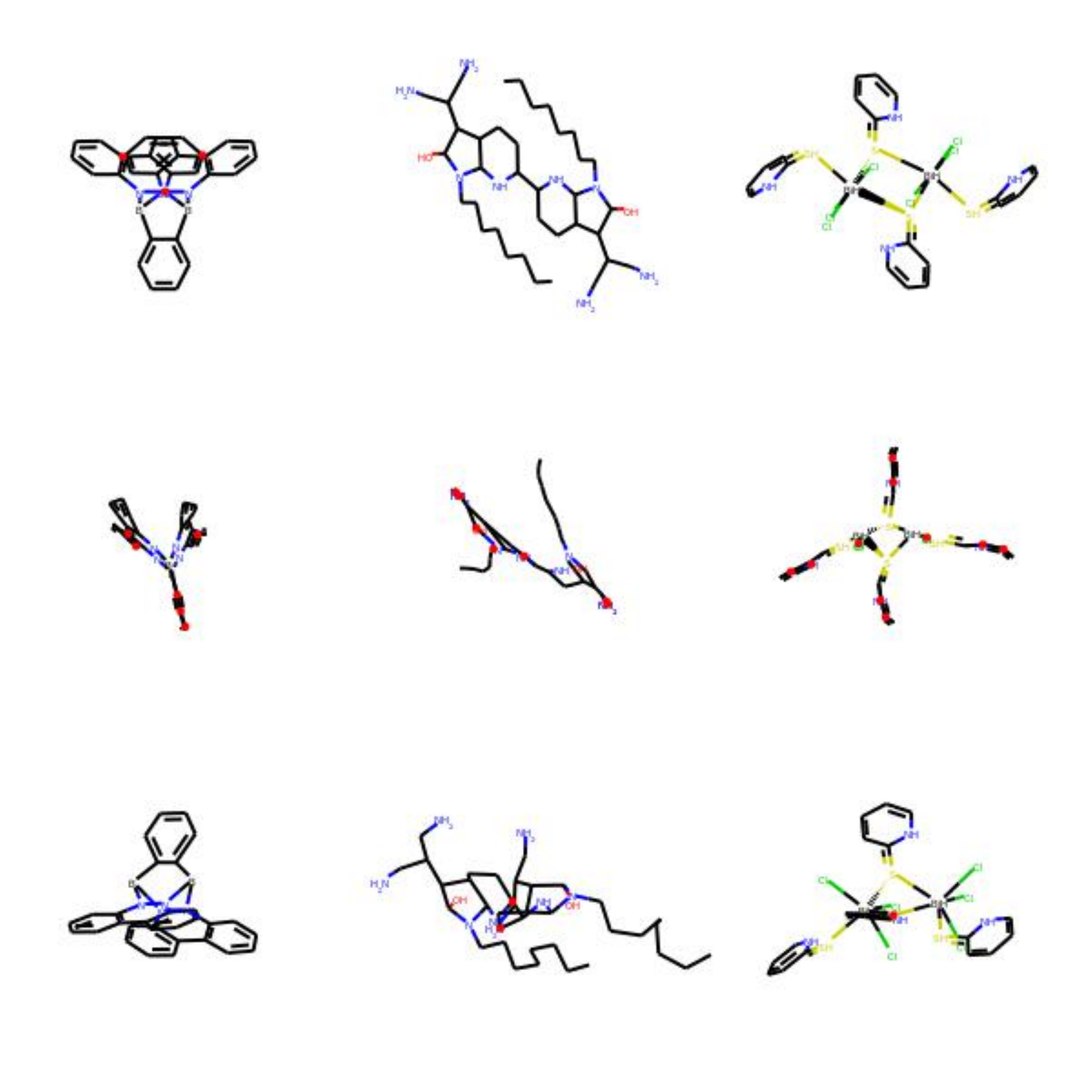}
        \caption{CSD greatest difference in favour of neural network predictions}
        \label{CSD best nnvff}
    \end{subfigure}
    \rulesep
    \begin{subfigure}{0.45\textwidth}
        \includegraphics[width=\textwidth,height=0.22\textheight]{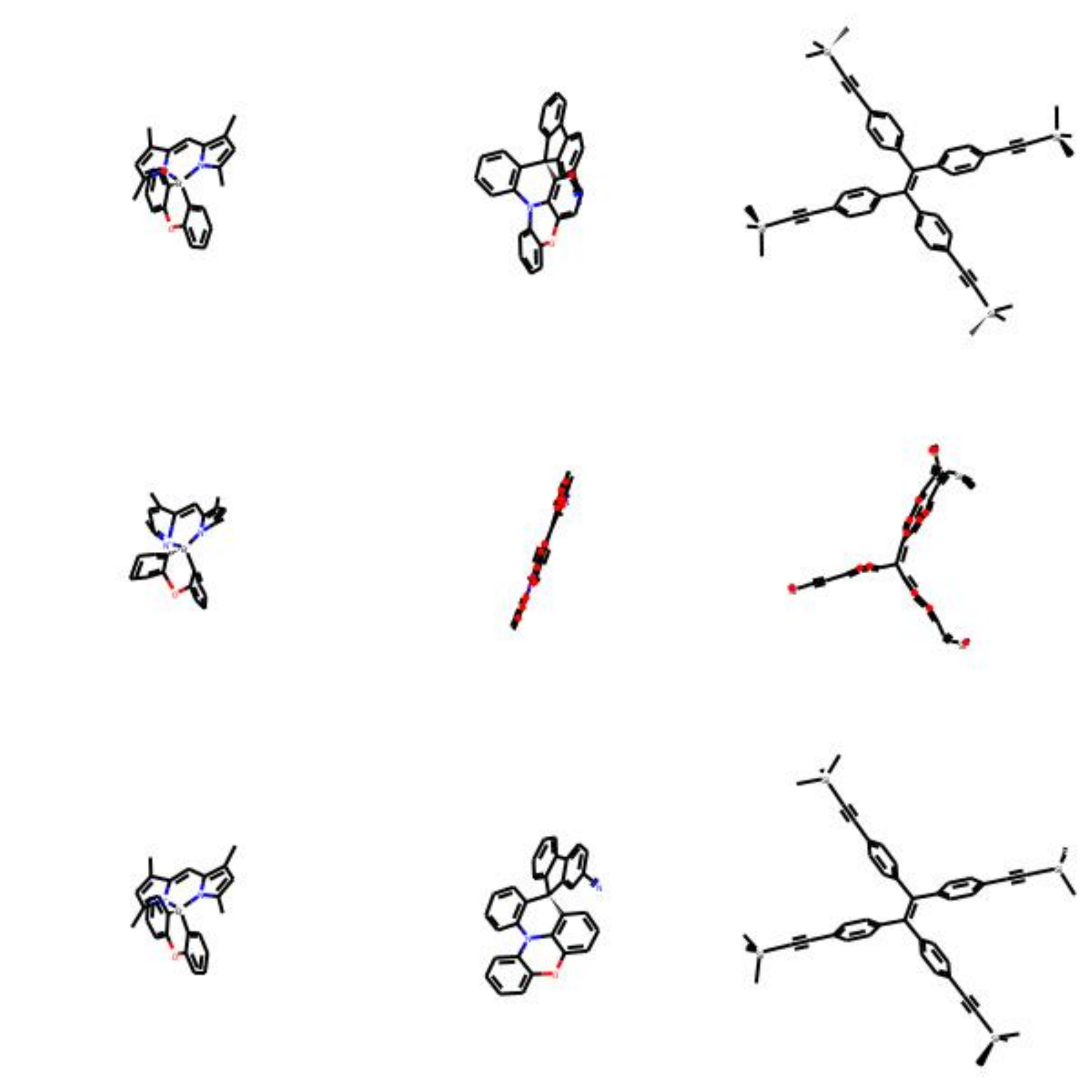}
        \caption{CSD greatest difference in favour of ETKDG + MMFF predictions}
        \label{CSD worst nnvff}
    \end{subfigure}
    \caption{This figure shows the three molecules in each dataset for which the differences between the RMSDs of the neural network predictions and the baseline ETKDG + MMFF predictions were greatest in favour of the neural network predictions ($max \ (RMSD_{CVGAE}-RMSD_{ETKDG+MMFF})$), and the three for which this difference was greatest in favour of the ETKDG + MMFF predictions ($max \ (RMSD_{ETKDG+MMFF}-RMSD_{CVGAE})$). The top row of each subfigure contains the reference molecules, the middle row contains the neural network predictions and the bottom row contains the conformations generated by applying MMFF to the reference conformations.}
    \label{nn_v_ff}
\end{figure}

\begin{figure}[htbp]
    \begin{subfigure}{0.45\textwidth}
        \includegraphics[width=\textwidth,height=0.22\textheight]{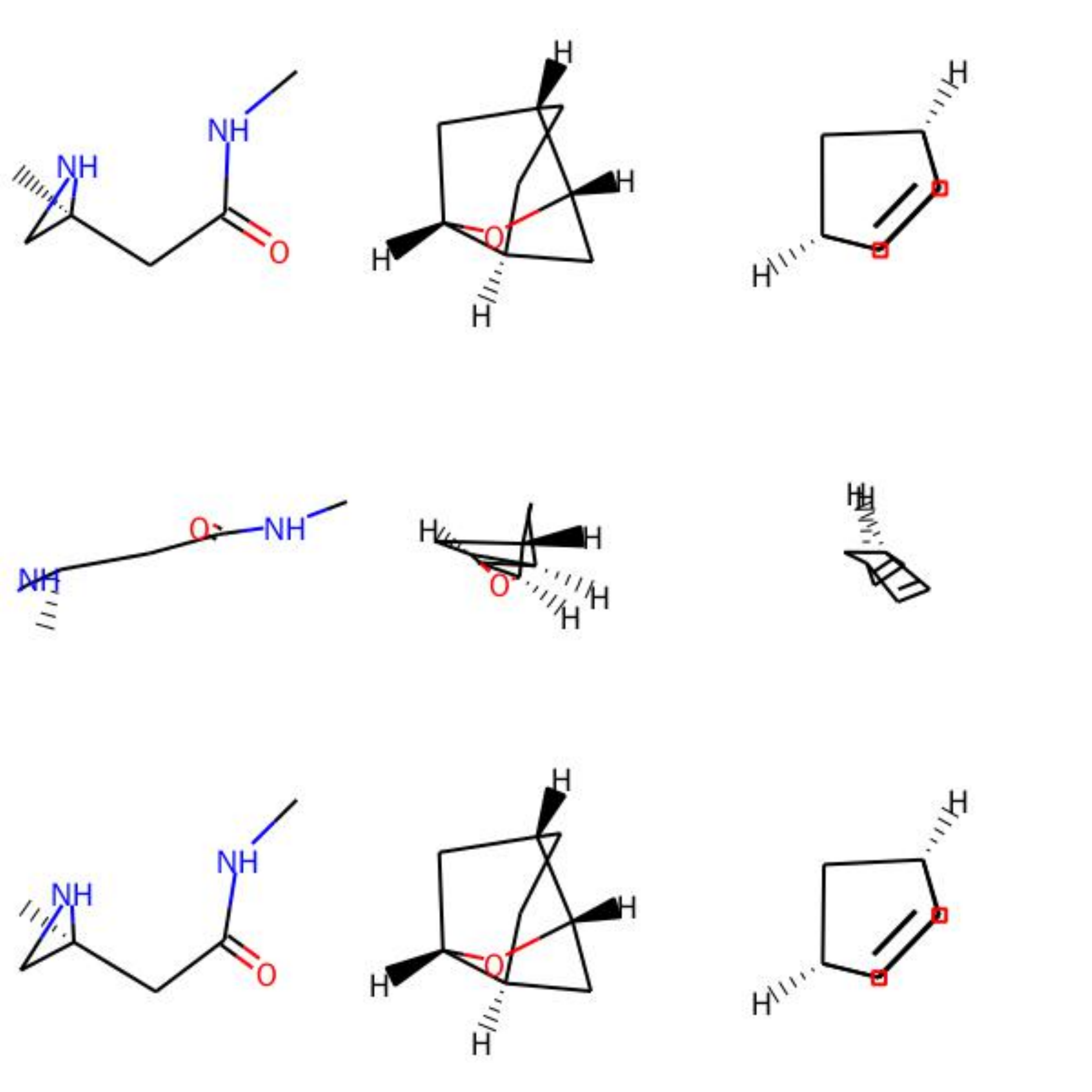}
        \caption{QM9 greatest improvement}
        \label{QM9 best nnffvnn}
    \end{subfigure}
    \rulesep
    \begin{subfigure}{0.45\textwidth}
        \includegraphics[width=\textwidth,height=0.24\textheight]{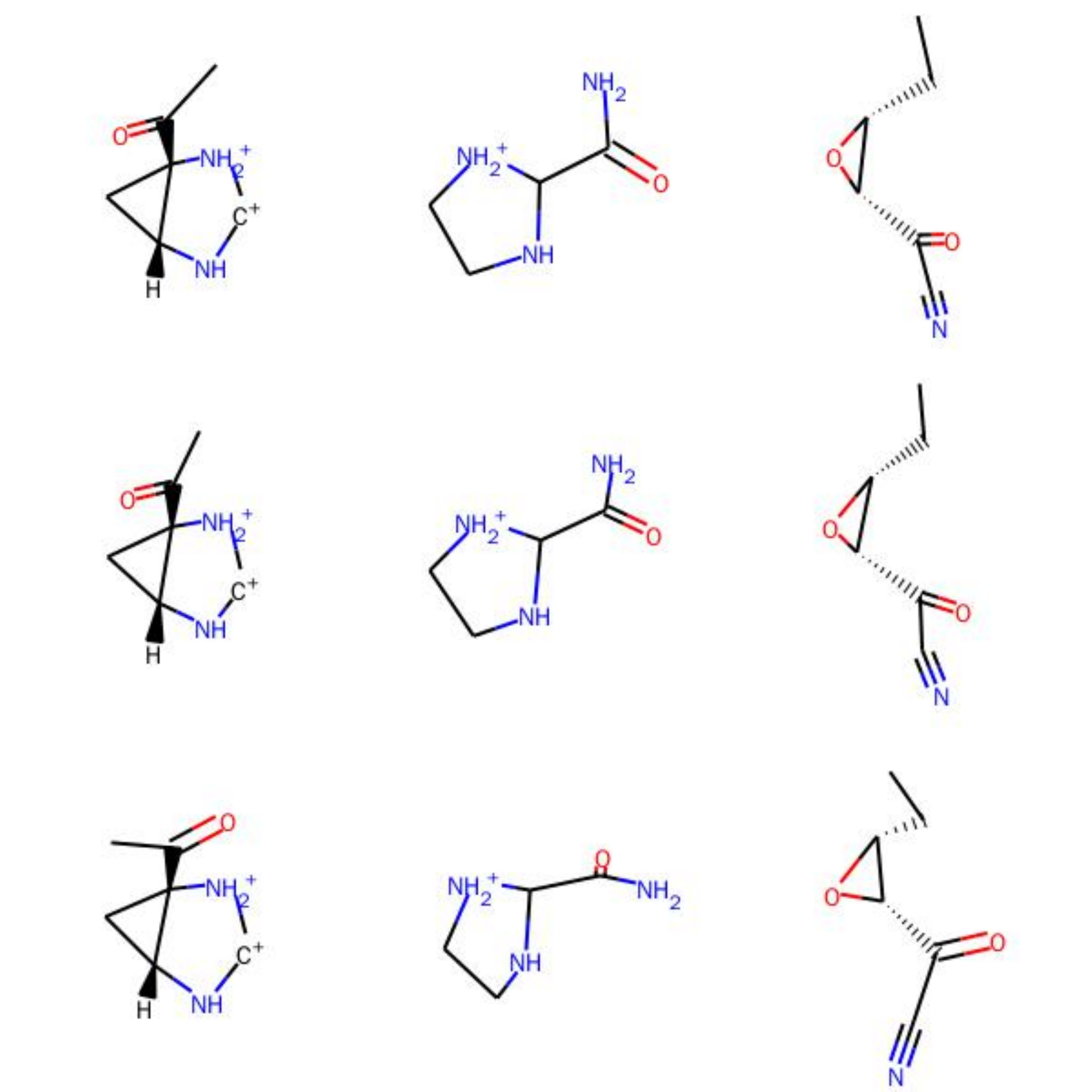}
        \caption{QM9 greatest deterioration}
        \label{QM9 worst nnffvnn}
    \end{subfigure}
    \hrule
    \begin{subfigure}{0.45\textwidth}
        \includegraphics[width=\textwidth,height=0.25\textheight]{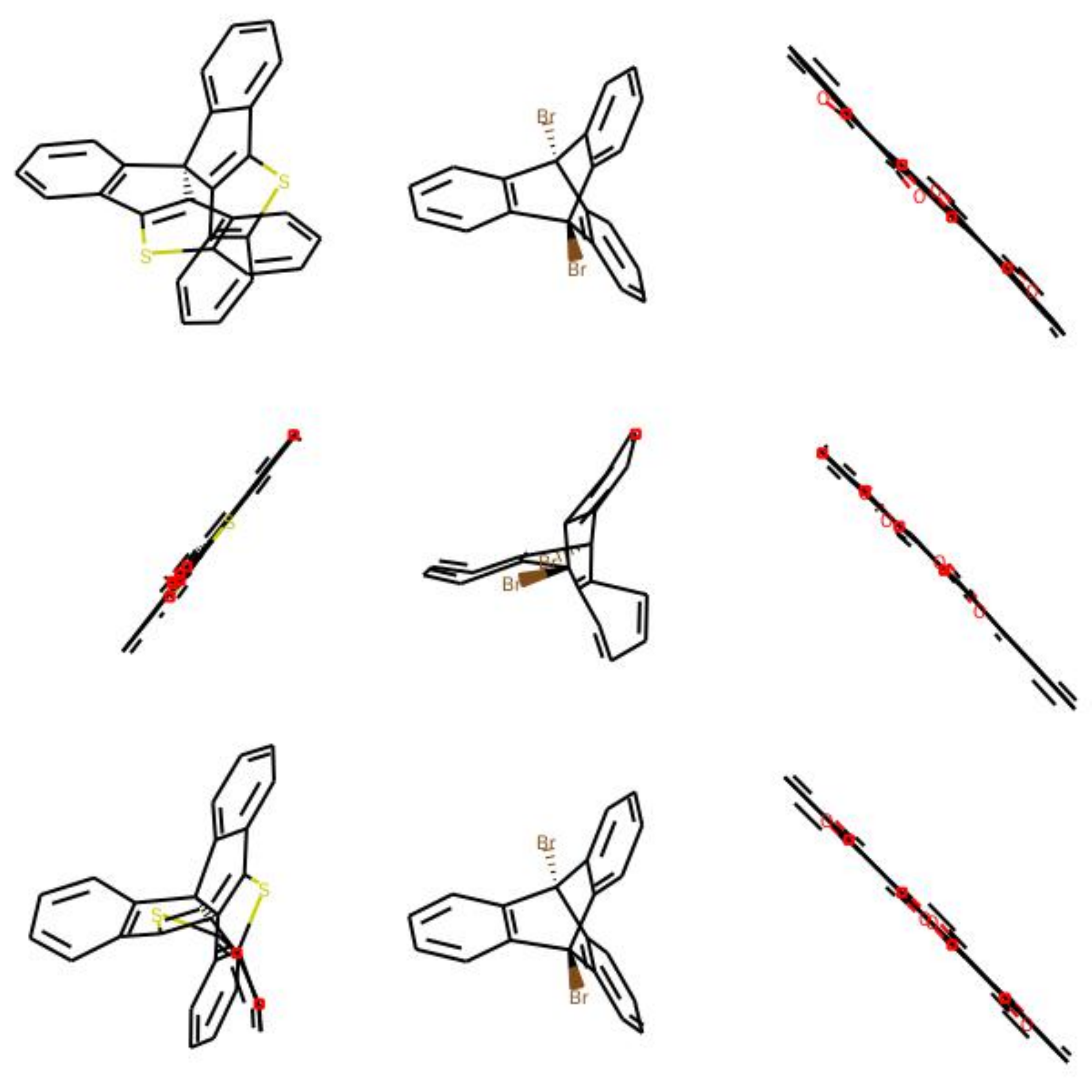}
        \caption{COD greatest improvement}
        \label{COD best nnffvnn}
    \end{subfigure}
    \rulesep
    \begin{subfigure}{0.45\textwidth}
        \includegraphics[width=\textwidth,height=0.25\textheight]{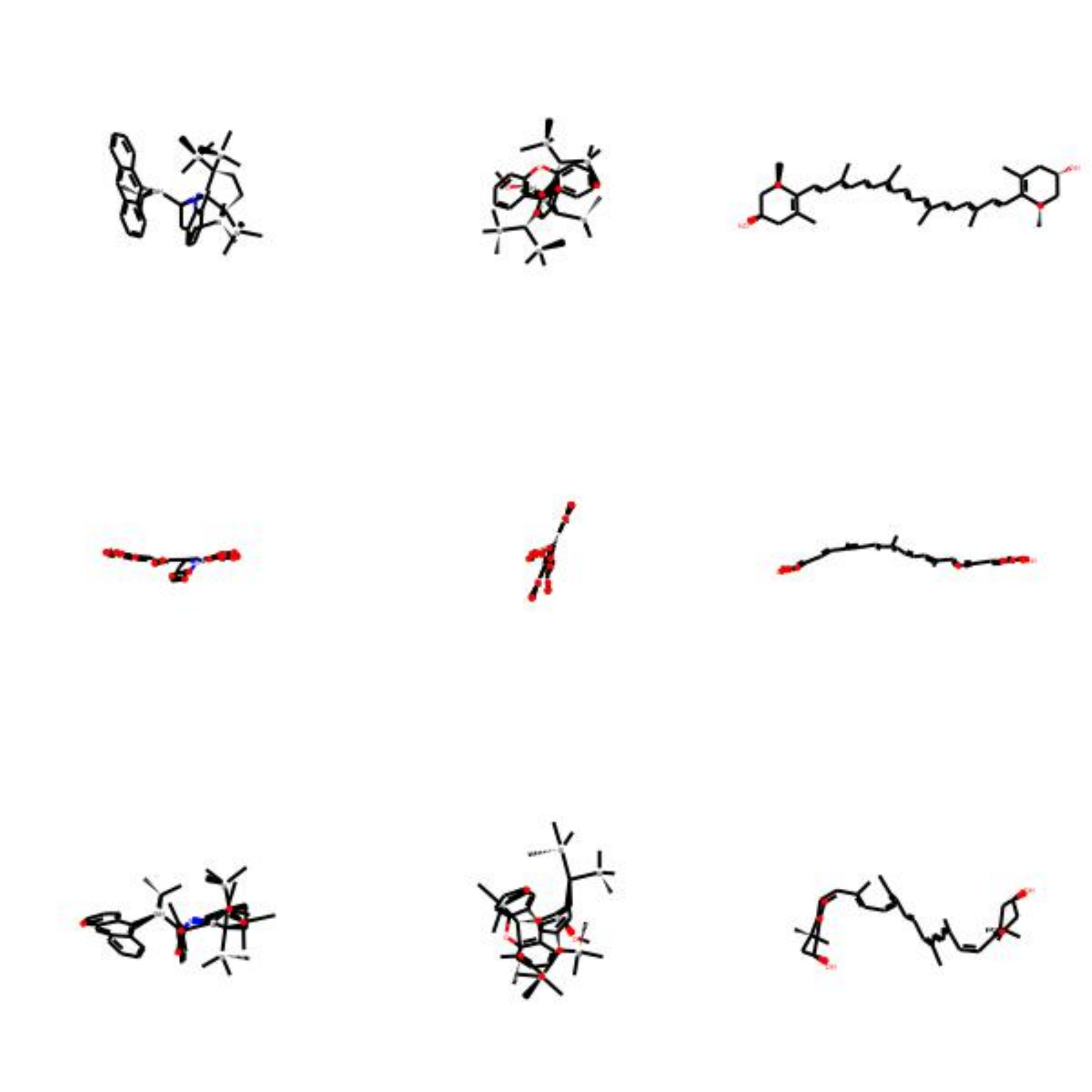}
        \caption{COD greatest deterioration}
        \label{COD worst nnffvnn}
    \end{subfigure}
    \hrule
    \begin{subfigure}{0.45\textwidth}
        \includegraphics[width=\textwidth,height=0.25\textheight]{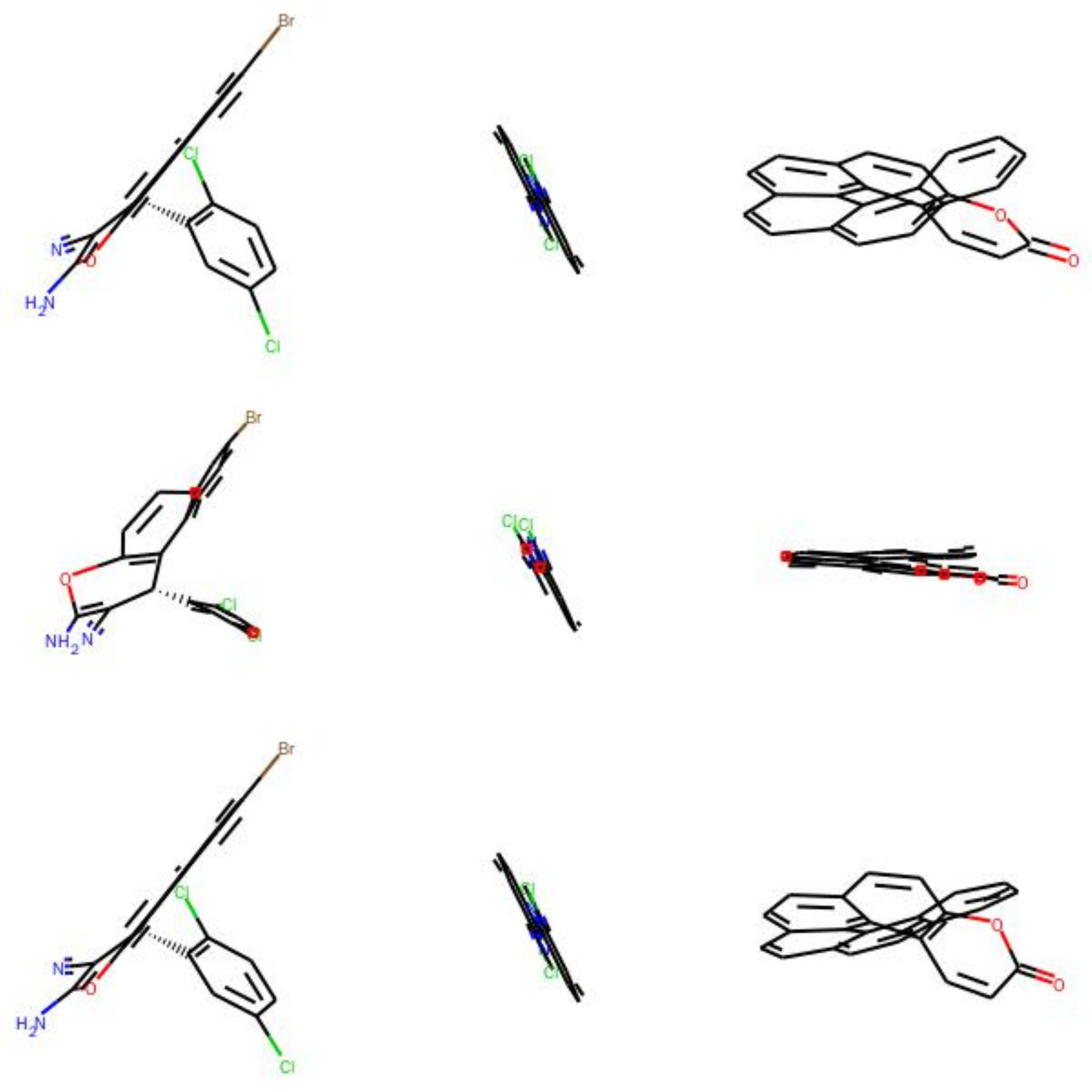}
        \caption{CSD greatest improvement}
        \label{CSD best nnffvnn}
    \end{subfigure}
    \rulesep
    \begin{subfigure}{0.45\textwidth}
        \includegraphics[width=\textwidth,height=0.25\textheight]{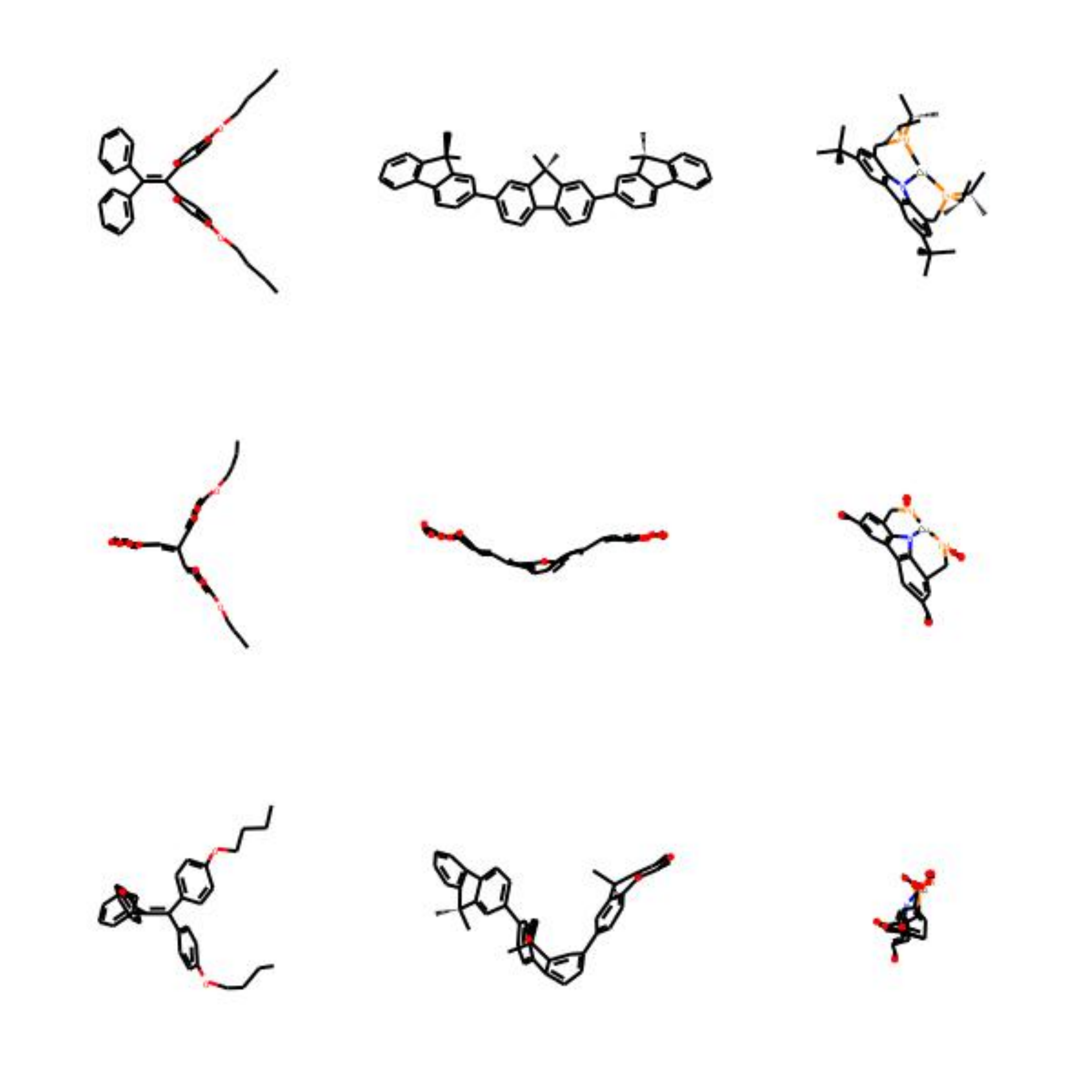}
        \caption{CSD greatest deterioration}
        \label{CSD worst nnffvnn}
    \end{subfigure}
    \caption{This figure shows the three molecules in each dataset whose RMSD decreased the most and the three whose RMSD increased the most on applying MMFF to the conformations predicted by the neural network. The top row of each subfigure contains the reference molecules, the middle row contains the neural network predictions and the bottom row contains the conformations generated by applying MMFF to the neural network predictions.}
    \label{nnff_v_nn}
\end{figure}

Figures \ref{nn_v_ff} and \ref{nnff_v_nn} qualitatively compare the results of CVGAE against MMFF and CVGAE + MMFF against CVGAE respectively. For each dataset, each figure shows the three molecules for which the first method in each figure outperforms the second method by the greatest amount, and the three molecules for which the second method outperforms the first by the greatest amount. The reference molecules are shown alongside the conformations resulting from each of the methods for comparison.

We can see some general trends from both these figures. The conformations produced by the neural network are qualitatively much more similar to the reference in the case of the QM9 dataset than in the cases of the COD and CSD datasets. In the case of the COD and CSD datasets, the CVGAE predictions appear to be squashed or compressed in comparison to the reference molecules. For example, in almost every case we can see the absence of visible rings and the absence of bonds protruding from the lengthwise dimension of the molecule. At the same time we can see that on COD and CSD, CVGAE does better than ETKDG + MMFF in cases where ETKDG + MMFF creates loops and protrusions in the wrong places.

\section*{Analysis and Future Work}

Overall we observe that CVGAE performs better than ETKDG + MMFF on QM9 than on COD and CSD. One possible reason that could explain this phenomenon is that COD and CSD contain much larger number of heavy atoms per molecule than QM9. In the absence of adequate number of neural message passing steps and adequate number of hidden units, the network may converge to outputting a conformation that contains atoms largely along a single non-linear dimension in order to minimize outliers, which would be heavily penalized by the sum of squared distances term in the loss function. A neural network architecture with a larger number of neural message passing steps and larger number of hidden units may be needed to generate less conservative conformations and achieve comparable results to those for QM9. This is a recommended direction of future work that will require more computational resources, including distributed training on multiple GPUs with sufficient memory.

Another concern for COD and CSD is the inconsistency in the environments from which the reference conformations are obtained. The inconsistency would not be a serious concern for small molecules, but it can result in performance degradation with larger molecules. Further investigation should be performed with the dataset of larger molecules and their reference conformations whose corresponding environments are identical. Additionally, conditioning deep generative graph neural networks on the environment could be explored in the future.
 
We also observe that our CVGAE method has a lower variance than the baseline methods, so a relatively small number of samples needs to be taken before getting a conformation with a good RMSD. In addition, CVGAE is faster than force field methods and uses less computational resources once trained. Using conformations generated by CVGAE as an initialization to force field method showed promising results on the QM9 dataset that allowed to combine the best of two distinct methods. However, applying a force field method on the conformations generated by CVGAE leads to an increase in RMSD on the COD and CSD datasets - future work could explore why this is the case. Another avenue of future inquiry could be the joint training of CVGAE and a force field method, which would involve implementing force field minimization using a deep learning framework, connecting this to CVGAE and backpropagating through this aggregate model. This joint training could further yield better results than either method alone.

\section*{Data availability}

The source code and preprocessed datasets are available online at {https://github.com/nyu-dl/dl4chem-geometry}.

\bibliography{manuscript}

\section*{Acknowledgements}

S.K. was supported by the National Research Foundation of Korea (NRF) grant funded by the Korea government (Ministry of Science and ICT) (No. NRF-2017R1C1B5075685). K.C. was partly supported by Samsung Research and thanks support by eBay, TenCent, NVIDIA and CIFAR.

\section*{Author contributions}

S.K. and K.C. have conceived the initial idea and started the project. E.M., O.M. and S.K. have run the experiments and further refined the idea of the project. E.M., O.M., S.K. and K.C. have written the paper.

\section*{Additional information}

\subsection*{Competing interests}
The authors declare no competing interests.

\end{document}